\documentclass[10pt,twocolumn,letterpaper]{article}

\usepackage{iccv}
\usepackage{times}
\usepackage{epsfig}
\usepackage{graphicx}
\usepackage{amsmath}
\usepackage{amssymb}

\usepackage[ruled,lined,vlined,linesnumbered]{algorithm2e}
\usepackage{multirow}

\usepackage[pagebackref=true,breaklinks=true,letterpaper=true,colorlinks,bookmarks=false]{hyperref}
 \normalsize

\iccvfinalcopy 

\def\iccvPaperID{1205} 

\newcommand*{\affaddr}[1]{#1} 
\newcommand*{\affmark}[1][*]{\textsuperscript{#1}}
\newcommand*{\email}[1]{\texttt{#1}}

\ificcvfinal\fi
\begin{document}

\title{Smart Mining for Deep Metric Learning}
\author{\thanks{Vijay Kumar B G and Ben Harwood contributed equally to this work}    Ben Harwood\affmark[2], \footnotemark[1]     Vijay Kumar B G\affmark[1], Gustavo Carneiro\affmark[1], Ian Reid\affmark[1], Tom Drummond\affmark[2]\\
\affaddr{\affmark[1]The University of Adelaide},\affaddr{\affmark[2]Monash University}\\
\email{\tt\small \{vijay.kumar,gustavo.carneiro,ian.reid\}@adelaide.edu.au},  \email{\tt\small \{ben.harwood,tom.drummond\}@monash.edu}\\}

\maketitle

\begin{abstract}
To solve deep metric learning problems and producing feature embeddings, current methodologies  will commonly use a triplet model to minimise the relative distance between samples from the same class and maximise the relative distance between samples from different classes.
Though successful, the training convergence of this triplet model can be compromised by the fact that the vast majority of the training samples will produce gradients with magnitudes that are close to zero. 
This issue has motivated the development of methods that explore the global structure of the embedding and other methods that explore hard negative/positive mining. The effectiveness of such mining methods is often associated with intractable computational requirements. In this paper, we propose a novel deep metric learning method that combines the triplet model and the global structure of the embedding space. We rely on a smart mining procedure that produces effective training samples for a low computational cost.
In addition, we propose an adaptive controller that automatically adjusts the smart mining hyper-parameters and speeds up the convergence of the training process.
We show empirically that our proposed method allows for fast and more accurate training of triplet ConvNets than other competing mining methods. Additionally, we show that our method achieves new state-of-the-art embedding results for CUB-200-2011 and Cars196 
datasets.
\end{abstract}

\section{Introduction}

\begin{figure*}
	\centering
	\includegraphics[width=0.9\linewidth]{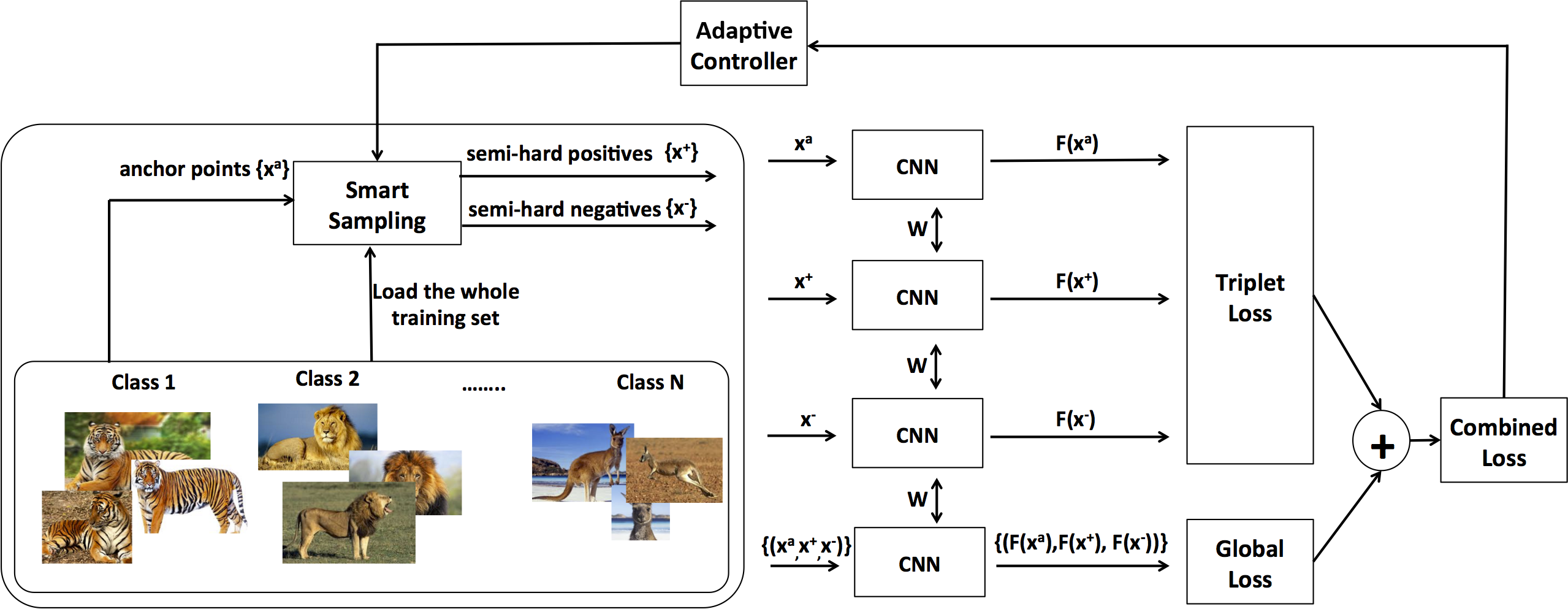} 
	\caption{Our proposed deep metric learning model that combines a triplet and a global loss and uses a smart sampling procedure that is capable of quickly searching the entire training set to select effective training samples.  The hyper-parameters of the smart sampling are automatically estimated by the proposed adaptive controller with the goal of accelerating the training process.}
	\label{fig:main}
\end{figure*}

The development of deep metric learning models for the estimation of effective feature embedding~\cite{dosovitskiy2014discriminative,han2015matchnet,kumar2016learning,masci2014descriptor,shrivastava2016training,simo2015discriminative,Simonyanpami14,song2016learnable,ustinova2016learning,wohlhartcvpr15,ZagoruykoCVPR15,yuhuihardaware2017,Nofuss_Movshovitz-Attias17,indefence_HermansBL17} is at the core of many recently proposed computer vision methods~\cite{guillaumin2010multiple,schroff2015facenet,song2016deep,wang2014learning,zhuang2016fast}.  The main advantage of such models lies in their ability to automatically learn metric spaces, where samples from similar classes tend to be close together, while samples from different classes are more likely to be far from each other.  The main scenario envisaged for such an approach involves the presence of an extremely large number of classes (more than $10^5$ classes) and low number of samples per class (in $[10^1,10^2]$), where the implementation of traditional classifiers becomes challenging~\cite{song2016deep,song2016learnable}.  

Arguably, the most explored deep learning model that can estimate feature embedding is based on triplet networks~\cite{hoffer2014deep,wang2014learning}, which are an extension of the siamese network~\cite{bromley1993signature}.  Triplet networks are composed of three identical convolutional neural networks (ConvNets) that are trained using triplets of samples: an anchor sample, a positive sample of the same class as the anchor, and a negative sample of a different class.  The training procedure is based on a loss function that penalises large relative distances between the anchor and positive samples and small relative distances between the anchor and negative samples.  Therefore, this training procedure relies on triplets that contain hard positive cases (anchor and positive samples that are far apart) and hard negative cases (anchor and negative samples that are close together).  In other words, these hard samples will form triplets that produce gradients with sufficiently large magnitude.  Assuming that a training set has $N$ samples, then the set of triplets has complexity size $O(N^3)$, which means that its formation is infeasible even for datasets of modest sizes (e.g., $N=10^5$).  This issue has lead to the implementation of importance sampling techniques~\cite{schroff2015facenet,simo2015discriminative,wang2014learning} that stochastically under-samples the set of triplets. Here, their success relies on using enough samples to guarantee that a certain fraction of the hard positives and negatives are available for training~\footnote{We have not found a formal study that describes the number of samples used for training versus the fraction of hard positive/negatives.}.  
Given the high complexity involved in finding hard positive and negative samples, another training procedure has been developed in order to guarantee training samples with large gradient magnitudes: the incorporation of loss functions that take into account the global structure of the embedding space~\cite{kumar2016learning,song2016learnable,ustinova2016learning}.

In this paper, we propose a novel deep metric learning approach that combines a global~\cite{kumar2016learning} and a triplet loss~\cite{hoffer2014deep,wang2014learning} computed using training samples acquired from a smart sampling method that has low computational complexity~\cite{harwood2016fanng} and can find effective training samples that produce gradients with large magnitude (see Fig.\ref{fig:main}).  Essentially, our smart sampling method circumvents the importance sampling issue mentioned above, enabling our model to be robustly trained with more effective hard negatives and positives, and without the need for a stochastic under-sampling of the training set.  
Additionally, we propose a novel adaptive controller that accelerates learning by monitoring training performance, estimating its own internal parameters and then automatically adjusting the smart sampling hyper-parameters.
We show empirically that our proposed method allows for fast and more accurate training of triplet ConvNets than other competing mining methods. Additionally, we show that our method achieves new state-of-the-art embedding results for CUB-200-2011 and Cars196 
datasets.


\section{Related Work}
\label{sec:related_work}

In this section, we review recent approaches for selecting hard positives and negatives for training triplet and siamese networks,
methods that explore the global structure of the embedding space,
and the approximate nearest neighbour search that forms the basis of our proposed method. As pointed out by Shrivastava et al.~\cite{shrivastava2016training}, hard negative and positive mining is a relabeling of the problem of \emph{bootstrappping}~\cite{sung1996learning},  where the idea is to start the training of the embedding model with triplets containing positives and negatives that appear to be well separated, and gradually introduce more challenging positive and negative samples as we progress with the training.  One of the major issues associated with this approach is on how to introduce such challenging samples - in particular: 1) how to effectively and efficiently sample the training set to select effective training samples, particularly considering that there are $N^3$ triplets from a training set containing $N$ samples, and 2) what is the definition of challenging positive and negative samples.

Wang et al.~\cite{wang2014learning} described a way to build triplets based on a manual annotation of sample relevance.  Using such relevance, the idea is to use importance sampling to build triplets, but this approach is limited by the fact that it needs those manual annotations.
More recently proposed approaches rely on image label, such as the siamese network that gradually introduces hardest possible positive and negative samples~\cite{simo2015discriminative}.  This is achieved by randomly sampling the training set for pairs of anchor and positive samples, and sorting these pairs in descending order with respect to the distance between the two samples in the embedding space.  A similar approach is applied for pairs of anchor and negative samples, but the sorting is in ascending order.  Then, the training pairs are formed by the top pairs in both lists.  We use this sampling scheme as the hard mining baseline.  Han et al.~\cite{han2015matchnet} introduced an efficient reservoir sampling method to select positive and negative samples, but they do not apply any type of importance sampling to select challenging samples.  In FaceNet, Schroff et al.~\cite{schroff2015facenet} introduced a triplet training approach, where pairs of anchor and positive samples are randomly selected, and pairs of anchor and negative samples are selected from a subset of the training set (i.e., the mini-batch in regular deep learning model training) using a criterion that selects "semi-hard" negatives: pairs of anchor and negative samples are selected if they are close, but at least farther than the anchor-positive pair.  This semi-hard negative sampling improves the robustness of training by avoiding overfitting outliers in the training set.  An efficient computation of the full matrix of pairwise distances of a subset of the training set (i.e., the mini-batch) allows Song et al.~\cite{song2016deep} to design of a new loss function that integrates all positive and negative samples to form a lifted structured embedding.  However, differently from our work, the lifted structured embedding only works for the mini-batch instead of the full training set.  

The aforementioned issues present in the training of triplet models has motivated the development of approaches that explore the global structure of the embedding space.
Kumar et al.~\cite{kumar2016learning} proposed a global loss function that uses first and second order statistics to allow for robust training of triplet networks in an approach that improves the training robustness, but still relies on stochastic sampling of positive and negative samples. Ustinova and Lempitsky~\cite{ustinova2016learning} presented a loss function that minimises the integral of the product between the distribution of negative similarities and the cumulative density function of the positive similarities.  
Similarly, Song et al.~\cite{song2016learnable} introduced a loss function that optimises a global clustering quality metric (NMI).
As shown by Kumar et al.~\cite{kumar2016learning}, it appears that a combination of local and global losses can produce the most effective embedding spaces, so we believe that the last two approaches mentioned above~\cite{ustinova2016learning,song2016learnable} still have room for improvement, but that improvement depends on more effective hard negative and positive sampling approaches.

In seeking a more effective approach to find hard triplets, we observe that hard negative mining (and to a lesser extent hard positive mining) can be framed as an instance of the well studied approximate nearest neighbour (ANN) search problem. In particular, when mining for negatives we are primarily interested in avoiding the computational cost of exhaustively searching through the entire training set. Fortunately, ANN search methods are able to trade-off a small decrease in nearest neighbour recall for large gains in computational efficiency. In the context of hard negative mining, a small set of nearest neighbours in the current embedding can be guaranteed to contain samples from at least two difference classes (due to training with very few samples per class). A FANNG (Fast Approximate Nearest Neighbour Graph)~\cite{harwood2016fanng} is a graph based index that can find these neighbourhoods quickly and with a very high rate of recall. Additionally, FANNGs are built in the full embedding space which allows the triplet selection to reuse exact distances that have been computed during the ANN search. FANNG provides state-of-the-art performance at high recall rates while adding only a single tuning parameter for the indexing quality and another for the ANN search quality.

\section{Proposed Method}
\label{sec:proposed_method}

We first describe the architecture of a triplet network~\cite{wang2014learning,hoffer2014deep,schroff2015facenet,wohlhartcvpr15} and the loss function used in its training.  Then, we describe the sampling method applied in the training process. 
Assume that the training set is represented by $\mathcal{T} = \{ (\mathbf{x}_i,y_i) \}_{i=1}^N$, with $\mathbf{x}_i \in \mathbb R^{n \times n}$ and $y_i \in \{1,...,C\}$.  The feature embedding is denoted by $f(\mathbf{x},\theta_f)$, where $f: \mathbb R^n \times \mathbb R^k \rightarrow \mathbb R^m$, with $\theta_f \in \mathbb R^k$ representing the network parameters (weight matrices, bias vectors, and normalisation parameters).   The triplet network comprises three identical deep convolutional neural networks (ConvNet) containing $L$ layers, each defined by:
\begin{equation}
f(\mathbf{x},\theta_f) = f_{\text{out}} \circ r_{L} \circ h_L \circ f_L \circ ... \circ r_1 \circ h_1 \circ f_1(\mathbf{x}),
\label{eq:def_CNN}
\end{equation}
where $\theta_f$ is defined above, $f_l(.)$ denotes the linear transforms, $h_l(.)$ represents a normalisation function, and $r_l(.)$ denotes a non-linear activation function (e.g., ReLU~\cite{nair2010rectified}).  Also in \eqref{eq:def_CNN}, note that 
$\mathbf{f}_l = [\mathbf{f}_{l,1},...,\mathbf{f}_{l,n_l}]$ represents an array of $n_l$ pre-activation functions.

\subsection{Triplet Networks}
\label{sec:triplet_networks}

The triplet network~\cite{wang2014learning,hoffer2014deep,schroff2015facenet,wohlhartcvpr15} is trained with 
an input composed of an anchor point $\mathbf{x}_i$ (from class $y_i$), another point from the same class $\mathbf{x}_i^+ = \mathbf{x}_j$ (with $i \neq j$ and $y_i = y_j)$, and a point from a different class $\mathbf{x}_i^- = \mathbf{x}_k$ ((with $k \neq i$ and $y_i \neq y_k$).  The loss function for each triplet is defined by:
\begin{equation}
\begin{split}
J^t  &  ( \mathbf{x}_i,\mathbf{x}_i^+, \mathbf{x}_i^-, \theta_f ) =  \\ 
& \max \left ( 0,1-\frac{ \| f^{(1)}(\mathbf{x}_i, \theta_f) - f^{(3)}(\mathbf{x}_i^-, \theta_f) \|_2 } { \| f^{(1)}(\mathbf{x}_i, \theta_f) - f^{(2)}(\mathbf{x}_i^+, \theta_f) \|_2 + m }  \right ), \\
\end{split}
\label{eq:loss_triplet_embedding}
\end{equation}
where $m$ is the margin, $\mathbf{x}_i^+$ and $\mathbf{x}_i$ belong to the same class, $\mathbf{x}_i^-$ and $\mathbf{x}_i$ are from different classes, and $f^{(1)}(.)$, $f^{(2)}(.)$ and $f^{(3)}(.)$ are constrained to be the same network parameterized by $\theta_f$.  

The training of the triplet network can be made more robust with the introduction of a loss that explores the global structure of the embedding~\cite{kumar2016learning}.  In particular, the triplet loss in \eqref{eq:loss_triplet_embedding} can be extended with a global loss that assumes that the distribution of distances between anchor and positive samples and anchor and negative samples follow a Gaussian distribution.  This global loss aims to: 1) minimise the variance of the two distributions, 2) minimise the mean value of the distances between anchor and positive samples, and 3) maximise the mean value of the distances between anchor and negative samples, as follows:
\begin{equation}
\begin{split}
 J^g( \{ \mathbf{x}_i \}_{i=1}^N ,& \{ \mathbf{x}_i^+ \}_{i=1}^N, \{ \mathbf{x}_i^-\}_{i=1}^N , \theta_f ) = \\ &(\sigma^{2+}+\sigma^{2-})+\lambda \max\big(0,\mu^+ - \mu^- + t\big),
\end{split}
\label{eq:loss_global}
\end{equation}
where $\mu^+ = \sum_{i=1}^N d_i^+/N,~~ \mu^- = \sum_{i=1}^N d_i^-/N $, ~~
$\sigma^{2+} = \sum_{i=1}^N(d_i^+ -\mu^+)^2/N,~~~   \sigma^{2-} = \sum_{i=1}^N(d_i^- - \mu^-)^2/N$, with $\mu^+$ and $\sigma^{2+}$ denoting the mean and variance of the matching pair distance distribution,  
$\mu^-$ and $\sigma^{2-}$ representing the mean and variance of the non-matching pair distance distribution, 
 $d_i^+ = \frac{\lVert f^{(1)}(\mathbf{x}_i,\theta_f) - f^{(2)}(\mathbf{x}_i^+,\theta_f) \rVert_2^2}{4}$, $d_i^- = \frac{\lVert f^{(1)}(\mathbf{x}_i,\theta_f) - f^{(3)}(\mathbf{x}_i^-,\theta_f) \rVert_2^2}{4}$, 
$\lambda$ is a term that balances the importance of each term, $t$ is the margin between the mean of the matching and non-matching distance distributions and
$N$ is the size of the training set.
Note in (\ref{eq:loss_global}), that we assume a triplet network (i.e., $f^{(1)}(.)$, $f^{(2)}(.)$ and $f^{(3)}(.)$ are the same network), where the  
squared Euclidean distances of the matching and non-matching pairs of the $i^{th}$ triplet are constrained to be $0 \leq d_i^+, d_i^- \leq 1$ because of the division by 4, and the normalisation layer 
enforces that the norm of the embedding is 1.

\subsection{Smart Mining}
\label{sec:smart_mining}

As discussed in Sec.~\ref{sec:related_work}, semi-hard mining has proved an effective method for training triplet networks~\cite{schroff2015facenet} with the primary aim of finding sets of triplets that will continue to progress the training of the network. Naively, this can be achieved by selecting triplets that provide the greatest violation of the triplet constraint. For instance, given an anchor $\mathbf{x}_i$, the hardest positive is defined as
\begin{equation}
\mathbf{x}_i^+ = \underset{(\mathbf{x}_j,y_j) \in \mathcal{T} , \mathbf{x}_j \neq \mathbf{x}_i, y_j = y_i}{\arg \max} \lVert f^{(1)}(\mathbf{x}_i,\theta_f) - f^{(2)}(\mathbf{x}_j,\theta_f) \rVert_2^2,
\label{eq:hard_positive_sampling}
\end{equation}
and the hardest negative as
\begin{equation}
\mathbf{x}_i^- = \underset{(\mathbf{x}_j,y_j) \in \mathcal{T} , \mathbf{x}_j \neq \mathbf{x}_i, y_j \neq y_i}{\arg \min} \lVert f^{(1)}(\mathbf{x}_i,\theta_f) - f^{(3)}(\mathbf{x}_j,\theta_f) \rVert_2^2.
\label{eq:hard_negative_sampling}
\end{equation}
In order to avoid the costly $\arg \max$ over the entire training set, semi-hard mining is instead commonly performed over the stochastic subset of samples used in each mini-batch~\cite{simo2015discriminative,schroff2015facenet}. This method has the additional advantage of avoiding repeated attempts at learning from the hardest triplets that may never improve from epoch to epoch.

We define a novel off-line mining strategy that consists of first finding a set of approximate nearest neighbours $\mathcal{S} \subset \mathcal{T}$. Then, for all triplets with anchor $\mathbf{x}_i$ the set of neighbours $\mathcal{S}_i$ is used to determine appropriate positive and negative samples. To avoid mining poorly structured regions of the embedding, we limit our selection of negative samples to only include negatives where there is at least one positive sample that is closer to the anchor than the negative is. Positive samples are then chosen to guarantee a non-zero response from the loss function \eqref{eq:loss_triplet_embedding}.

\begin{figure}
	\centering
	\includegraphics[width=0.7\linewidth]{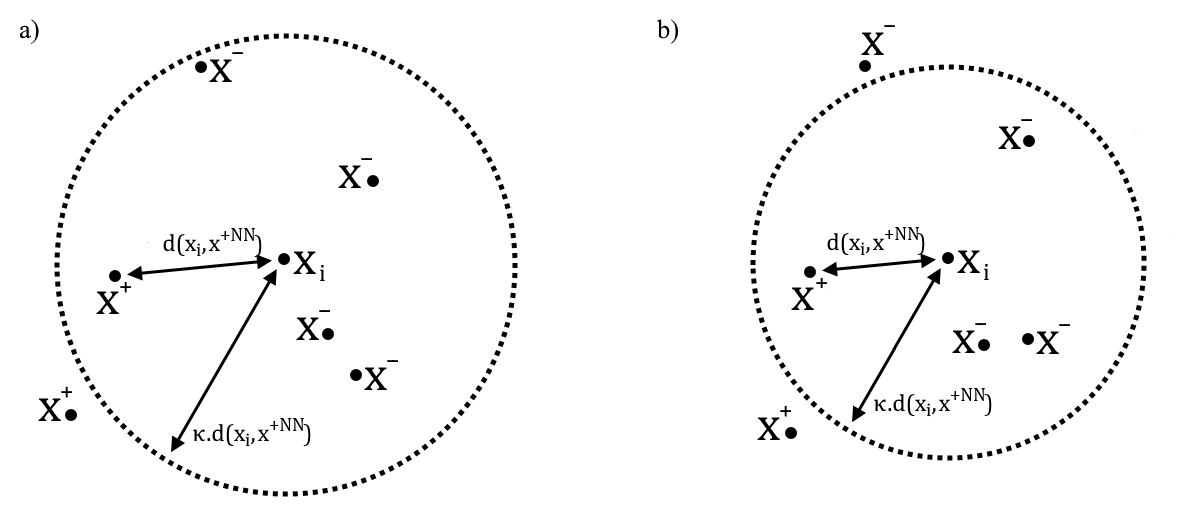} 
	\caption{A simplified 2-dimensional projection of the neighbours for anchor $\mathbf{x}_i$, here $\mathcal{S}_i$ contains two positive and four negative samples. Distance $d(x,y)$ is squared Euclidean. a) The current $\kappa$ and clustering of class $y_i$ specify an exclusion boundary containing all negatives samples, as such none are currently deemed suitable for training. b) On a subsequent epoch a smaller $\kappa$ and tighter class clustering now yeilds a negative sample outside the exclusion boundary. This negative and the positive sample further outside the exclusion boundary are used to complete a triplet that is guaranteed to violate the triplet constraint. }
	\label{fig:semi_hard_mining}
\end{figure}

More formally, we define a smart negative as any negative sample $\mathbf{x}_i^- \in \mathcal{S}_i$ such that
\begin{equation}
\begin{split}
\lVert f^{(1)}(\mathbf{x}_i,\theta_f) - &  f^{(3)}(\mathbf{x}_i^-,\theta_f) \rVert_2^2 \: > \\
\kappa & \cdot \lVert f^{(1)}(\mathbf{x}_i,\theta_f) - f^{(2)}(\mathbf{x}_i^{+NN},\theta_f) \rVert_2^2,
\end{split}
\label{eq:semi_hard_negative_sampling}
\end{equation} where $\kappa$ is a global tuning variable and $\mathbf{x}_i^{+NN}$ is the closest positive to $\mathbf{x}_i$ (note that this is not the positive used to form the triplet). The relationship between the exclusion boundary, the anchor, positive samples and negative samples can be seen in Figure \ref{fig:semi_hard_mining}.

Mining outside the region defined by the distance between an anchor $\mathbf{x}_i$ and the closest positive sample $\mathbf{x}_i^{+NN}$ ties the selection of negatives to how tightly the class $y_i$ is currently clustered in the embedding space. Additionally, the global parameter $\kappa$ provides a tunable scaling factor for the radius of these hyper-spherical exclusion boundaries centred on each anchor. Experimentally we have found that the best results are achieved by beginning training with a larger value for $\kappa$ and then gradually relaxing this constraint throughout the training. This allows previously excluded negatives to be selected for training during later epochs either because the positive neighbours have formed a tighter neighbourhood or the global exclusion value has been sufficiently reduced. The practical details of implementing this mining scheme are discussed below.

\subsubsection{Implementing Smart Mining with FANNG}

At the beginning of each training epoch we perform a full forward pass of the training set $\mathcal{T}$ to generate the current feature embedding $f(\mathbf{x},\theta_f)$. The indexing graph used in FANNG~\cite{harwood2016fanng} is then constructed using the traverse-add algorithm (Alg. 4 in ~\cite{harwood2016fanng}) with the embedding of each element of $\mathcal{T}$ forming a vertex in the graph. At each vertex, a list of outbound edges connect to un-occluded neighbours in a way that approximates the local surface structures of a lower dimensional manifold. Experimental results show that the order of these edge lists remains low (between 15-30 edges) and is independent of the size of the data set and the extrinsic dimensionality of the embedding space. The newly formed traversable graph enables the computationally efficient collection of the approximate nearest neighbour set $\mathcal{S}$.

As described in~\cite{harwood2016fanng}, the traverse-add algorithm can be repeatedly applied until a specified percentage success rate is reached. Once our target build percentage of 98\% is achieved, our approach diverges from the original building process for FANNGs. Rather than applying the backtrack search (Alg. 3 in ~\cite{harwood2016fanng}) to further refine the graph, we instead use the same backtrack search algorithm to immediately generate the approximate nearest neighbour set $\mathcal{S}$. Since the graph vertices provide a complete index of the training samples, we can compute each neighbour list $\mathcal{S}_i$ by passing the vertex $f(\mathbf{x}_i,\theta_f)$ to the backtracking search algorithm as both the query vector and starting point for the search. Because the collection of these neighbour lists does not modify the indexing graph, the searches can be performed in parallel. Each query returns a pre-specified number of nearest neighbours sorted in ascending order by distance from the query vertex, as well as the distances themselves. The size of the neighbour lists is selected to guarantee that both positive and negative samples will always be seen in the list.



\subsubsection{Triplet Construction}

\begin{algorithm}
	\DontPrintSemicolon
	\KwIn{Training samples $X$, Nearest neighbours $S$, Class labels $y$, Scale parameter $\kappa$}
	\KwOut{Mined triplets $T$}
	\For{each sorted neighbour list $s_i$}{
    	$neg \gets$ empty list of negatives\;
    	$pos \gets$ empty list of positives/valid negative range\;
		\For{each neighbour $s_{i}[j]$ of sample $x_i$}{
			\If{isEmpty($pos$)} {
            	\If{class($s_{i}[j]$) $\neq y_i$} {
               		\bf{continue}\;
                }
                $bound \gets \kappa \ \cdot$ distance($x_i, s_{i}[j]$)\;
            	$pos$.add($s_{i}[j], \emptyset$)\;
                \bf{continue}\;
        	}
            \If{distance($x_i, s_{i}[j]$) $< bound$} {
            	\bf{continue}\;
            }
            \If{class($s_{i}[j]$) $\neq y_i$} {
            	$neg$.add($s_{i}[j]$)\;
            }
            \If{class($s_{i}[j]$) $= y_i$} {
            	$pos$.add($s_{i}[j]$, clone($neg$))\;
            }
        }
        \For{each triplet $t[j]$ with anchor $x_i$}{
        	\If{isEmpty($neg$)} {
            	$t[j] \gets$ random triplet\;
                \bf{continue}\;
            }
            $t[j] \gets x_i, neg[0]$, random positive $\notin pos$\;
        	\For{each positive $pos[k]$}{
        		\If{$neg[0] \notin$ validRange($pos[k]$)} {
                	\bf{continue}\;
                }
            	$t[j] \gets x_i, neg[0], pos[k]$\;
                \bf{break}\;
            }
            $neg$.remove($neg[0]$)\;
        }
	}
	\bf{return} $T$\;
	\caption{Triplet Selection}
	\label{alg:triplet_selection}
\end{algorithm}

Once $\mathcal{S}$ is computed, the class label information $y$ is used to separate the neighbours into several lists. We perform a single iterative pass over each neighbour list $\mathcal{S}_i$ while maintaining a count of samples from class $y_i$ and a count of all samples from outside that class. Once the first positive sample has been found, the exclusion boundary is computed. Then any future negative samples that satisfy \eqref{eq:semi_hard_negative_sampling} are added to the list of valid negatives. Each subsequent positive sample is added to the list of valid positives along with the current number of valid negatives. With this information we can ensure that a positive sample is not put into a triplet with a later negative sample that is further from the anchor. Lastly, to construct each mined triplet in the current epoch, we take the first unused negative from the list of valid negatives associated with the triplets anchor as well as the first valid positive that is also valid for the chosen negative. Random triplets are used in rare cases where there are no valid negatives. If there are no valid positive samples associated with the chosen negative, then a positive sample is uniformly selected from the set $\mathcal{T} \backslash \mathcal{N}_i$. Algorithm \ref{alg:triplet_selection} illustrates this triplet selection process in pseudo-code. It is important to note that while each negative is used no more than once for a given anchor during any given epoch, positive samples can be used multiple times with the same anchor. However, the unique negatives will always ensure that no triplets are repeated. In general, our method will select softer negative and positive samples ahead of harder options.

\subsubsection{Runtime Complexity}

A naive hard mining algorithm that selects $O(N)$ triplets will have a worst case complexity of $O(N^3)$ on any given epoch. Assuming that the samples are equally distributed between C classes, then the complexity can instead be expressed as $\displaystyle O(N\frac{N}{C}\left(N-\frac{N}{C}\right))$. As $C \to N$, this complexity reduces to a best case of $O(N^2)$.

The smart mining algorithm requires the construction of a nearest neighbour index. Exhaustive index construction is $O(N^3)$ due to the sorting of all $N^2$ pairwise distances. However, we can guarentee a worst case complexity of $O(N^2)$ by instead building an approximation of the index. Using this index to find negatives up to the closest positive sample for each anchor can be performed with worst case complexity $O(N^2)$ regardless of class distribution. Given that $O(N^2)$ is the best case complexity for the naive hard mining approach above, we can conclude that our method is computationally more efficient.

For semi-hard mining, such as \cite{schroff2015facenet}, algorithmic complexity is reduced by limiting triplet selection to a brute force search within each minibatch. Given an epoch with M minibatches, the argmax for each anchor results in a total complexity of $\displaystyle O(M\left(\frac{N}{M}\right)^2)$, or simply $\displaystyle O(\frac{N^2}{M})$. For comparitive purposes, we note that larger minibatches (i.e. smaller M) tend to reduce training error \cite{simo2015discriminative} up until performance begins to be limited by the naive use of argmax. Even so, as $M \to 1$ the semi-hard mining complexity approaches $O(N^2)$ and the information available in each minibatch also approches that of both the naive and smart mining.

\subsubsection{Automatic Parameter Selection}

Up to this point, running our mining scheme requires hand tuning for the hyper-parameter $\kappa$. We propose a more robust solution that closes the loop on the triplet mining and training losses. At the beginning of each epoch, we would like to estimate what value of $\kappa$ will produce triplets of a suitable difficulty for the current network. One such goal could be to ensure that the error from the training set is consistent with the current error of the validation set. We estimate $\kappa$ with a simple linear model
\begin{equation}
\boldsymbol{\kappa} = \alpha \mathbf{e} + \beta,
\label{eq:controller_model}
\end{equation} that finds the least-squares solution for internal parameters $\alpha$ and $\beta$ from a vector of recent training errors $\mathbf{e}$ and their associated $\kappa$. Once we have computed the internal parameters, we can obtain the estimated value
\begin{equation}
\kappa = \alpha e_t + \beta,
\label{eq:controller_estimate}
\end{equation} by providing the current target error $e_t$. The model is initialised at the beginning of the third training epoch with an initial estimate of the internal parameters. At the beginning of the triplet mining on each subsequent epoch, the training results from the previous epoch are used to update the model. The inclusion of as little as 2\% mined triplets per batch is enough to control the training losses.


\begin{figure}
	\centering
	\includegraphics[width=0.7\linewidth]{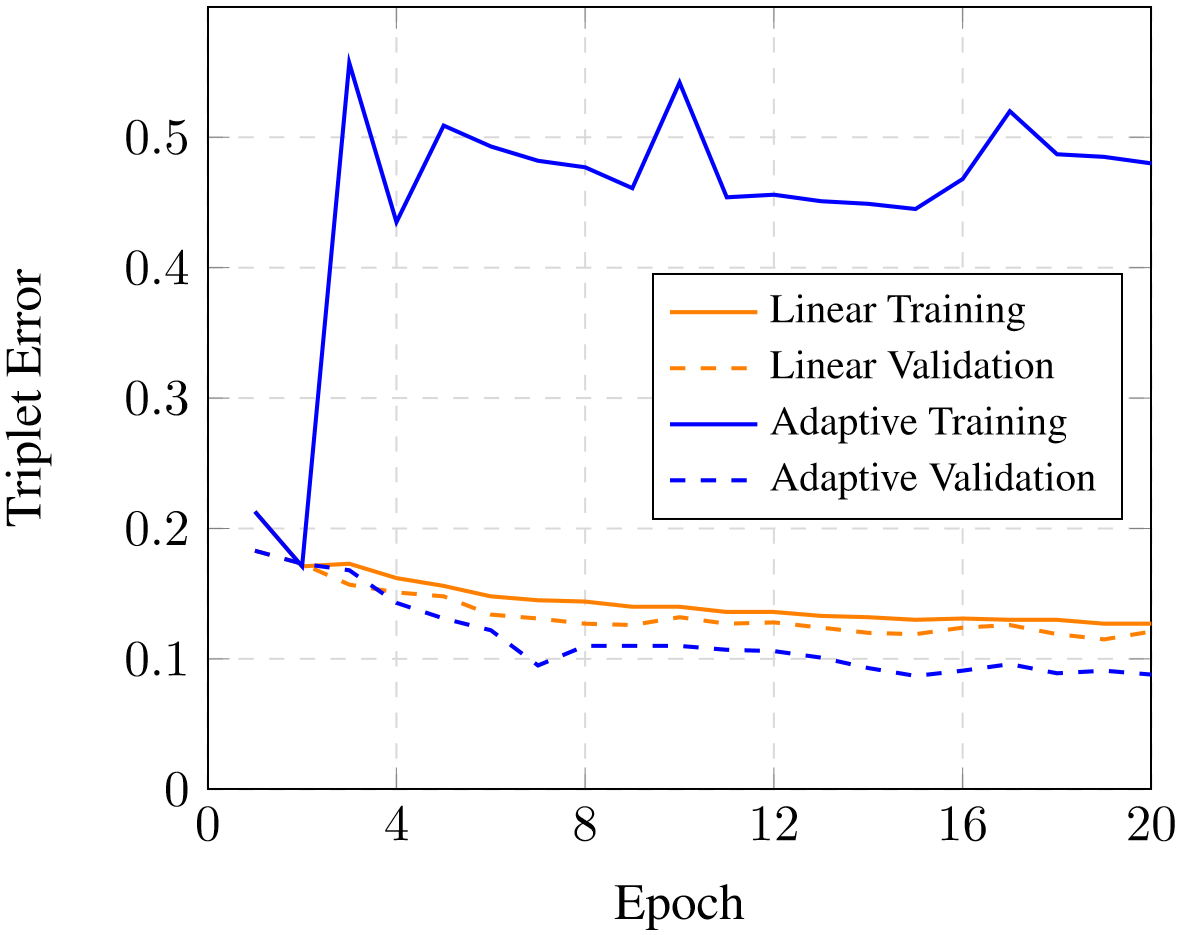} 
	\caption{ A comparison of training performance using hand tuned and adaptive selection of $\kappa$. Training and validation error is shown for the first 20 epochs while training on CUB-
200-2011~\cite{wah2011caltech}. }
	\label{fig:adaptive_error}
\end{figure}

As training progresses and the embedding improves, it is expected that both the training and validation error will decrease. Targeting a low training error will guarantee that most of the next epoch is spent on triplets that will not make a significant impact on the training. So instead, we can deliberately separate the training and validation errors so that the training error is kept high, while the validation error continues to decrease. To achieve this, we replace the use of the current validation error with a constant value that represents our target training error. Experimental results have shown that a target of between 50\% and 75\% training error is capable of producing more accurate embeddings in far fewer epochs. To maintain a high training error, it is best to use batches that are 50\% to 100\% mined triplets.

A comparison of hand tuned and adaptive parameter selection can be seen in figure \ref{fig:adaptive_error}. Training error gives an indication of the fraction of each batch that is producing a non-zero gradient and so can continue to shape the embedding space. The validation error is produced by evaluating the embedding with a reserved set of samples not used for training and is used as an inverse measure of the current quality of the embedding. Since the adaptive method is able to select harder triplets, while avoiding triplets that are so hard that the embedding structure could be damaged, we see that it can produce a higher quality embedding. Additionally, the steeper decent of the adaptive validation indicates that these results can be reached while also using fewer training epochs. In practice when using GPU accelerated code, our triplet selection accounts for less than 1\% of the total epoch runtime (the majority of the cost being the forward and backward propagation of the selected triplets). As such, the ability to produce high quality embeddings while converging in comparatively fewer epochs will greatly reduce the overall training time.

\section{Experiments}

For the experiments, we follow the protocol used in previous papers~\cite{sohn2016improved,song2016deep,song2016learnable}, which uses unseen classes from the CUB-
200-2011~\cite{wah2011caltech} and Cars196~\cite{krause20133d}
datasets
in order to
assess the clustering quality and k nearest neighbour retrieval~\cite{jegou2011product}.
Our proposed method combining {\bf triplet} and {\bf global} losses using {\bf FANNG}~\cite{harwood2016fanng} with and without automated hyper-parameter selection (i.e., the {\bf adaptive} controller) is compared with the following state-of-the-art deep metric learning approaches: (1) {\bf triplet}
learning with {\bf semi-hard} negative mining~\cite{schroff2015facenet} (with and without FANNG~\cite{harwood2016fanng}), (2)
{\bf lifted structured }embedding~\cite{song2016deep}, (3) {\bf N-pairs} metric loss~\cite{sohn2016improved}, (4) {\bf clustering}~\cite{song2016learnable}, and (5) {\bf triplet} combined with {\bf global} loss~\cite{kumar2016learning}.
For the approaches (1), (2), (3) and (4) above, we report the results from Song et al.'s paper~\cite{song2016learnable}.
For the remaining approaches (i.e. our proposed method, and (5) ), we use the same training and test set split described in~\cite{song2016deep} across all datasets. 
Specifically, the means CUB200-2011~\cite{wah2011caltech} has $11,788$ images of $200$ bird species, from which we take the first $100$ species for training and use the remaining $100$ species for testing. 
Cars196~\cite{krause20133d} has $16,185$ images from $196$ car models, from which we take the first $98$ classes for training and use the remaining $98$ for testing. For all our experiments, we initialize the network
with pre-trained GoogLeNet \cite{Szegedy_2015_CVPR} weights and randomly initialize
the final fully connected layer similar to \cite{song2016deep}. We set the embedding size to 64 \cite{song2016deep} and the
learning rate for the randomly initialized fully connected
layer is multiplied by 10 to achieve faster convergence similar
to \cite{song2016deep}.

For the experiments using triplet combined with global loss~\cite{kumar2016learning} and for our proposed approach, we let the training procedure run for a maximum of 20 epochs or until convergence (if fewer epochs were required). During the first two epochs, triplet mining was completely disabled to allow for batches comprised of only random triplets. Similar to \cite{schroff2015facenet,kumar2016learning}, we set the margin for the triplet and global loss to $0.2$ and $0.01$ respectively. We start experiment with an initial learning rate of $0.1$ and gradually decrease it by a factor of 2 after every 3 epochs. We use a weight decay of $0.0005$ for all of our experiments.

\subsection{Quantitative Results}

Here we report quantitative results based on the normalised mutual information (NMI)~\cite{manning2008introduction} score defined by the ratio of mutual information and product of entropies for two clustering assignments - this measures the label agreement between these two clustering assignments (ignoring the permutations). We also report the k nearest neighbour performance using the Recall@K metric~\cite{song2016learnable}.

Tables~\ref{table:results_cub} and \ref{table:results_cars}
show the NMI and k nearest neighbour performance
with the Recall@K metric results defined above comparing our method to the state of the art for the datasets CUB-
200-2011~\cite{wah2011caltech} and Cars196~\cite{krause20133d}.
From these tables, we can first see that {\bf Triplet + FANNG} significantly improves upon the {\bf Semi-hard}~\cite{schroff2015facenet} results with respect to all measures, and showing that the smart mining process using FANNGs is more effective than the more commonly used stochastic under-sampling of the training set.  
The combination of {\bf Triplet + FANNG + Global} shows gains with respect to all measures, when compared to {\bf Triplet + Global} and {\bf Triplet + FANNG}, demonstrating the importance of both the smart mining process and the use of a global loss.
The final model {\bf Triplet + FANNG + Global + Adaptive} shows competitive results with respect to all measures as well as a much faster convergence rate (see Fig.\ref{fig:recall_vs_epochs}). For instance, for the CUB-200-2011 dataset~\cite{wah2011caltech}, {\bf Triplet + FANNG + Global + Adaptive} converges in just four epochs, while {\bf Triplet + FANNG + Global} takes 20 epochs to converge. 
Similarly for Cars196~\cite{krause20133d}, {\bf Triplet + FANNG + Global + Adaptive} converges in just four epochs, while {\bf Triplet + FANNG + Global} takes 20 epochs to converge.
The accelerated rate of convergence is only achievable when the difficulty of the mined triplets is targeted at the right level for each individual epoch.


{\setlength{\tabcolsep}{0.45em}\begin{table}
    \caption{Clustering and recall performance on CUB-200-2011~\cite{wah2011caltech}. Our proposals are highlighted.}
	\label{table:results_cub}
	\begin{tabular}{llllll}
		\hline \noalign{\smallskip}
		{Method} & NMI & R@1 & R@2  & R@4 & R@8 \\
		\hline \noalign{\smallskip}
		Semi-hard~\cite{schroff2015facenet} & 55.38 & 42.59 & 55.03 & 66.44 & 77.23 \\[0.25em]
        Lifted Structure~\cite{song2016deep} & 56.50 & 43.57 & 56.55 & 68.59 & 79.63 \\[0.25em]
		N-pairs~\cite{sohn2016improved} & 57.24 & 45.37 & 58.41 & 69.51 & 79.49 \\[0.25em]
		Triplet + Global~\cite{kumar2016learning} & 58.61 & 49.04 & 60.97 & 72.33 & 81.85 \\[0.25em]
		Clustering~\cite{song2016learnable} & 59.23 & 48.18 & 61.44 & 71.83 &81.92 \\[0.25em]
		\bf{Triplet + FANNG} & 58.10 & 45.90 & 57.65 & 69.63 & 79.83 \\[0.25em]
        \bf{Triplet + FANNG +} \\
        \multicolumn{1}{r}{\bf{Global}} & \multirow{-2}{*}{\bf{60.09}} & \multirow{-2}{*}{49.44} & \multirow{-2}{*}{61.60} & \multirow{-2}{*}{73.09} & \multirow{-2}{*}{82.85} \\
		\bf{Triplet + FANNG +} \\
		\multicolumn{1}{r}{\bf{Global + Adaptive}} & \multirow{-2}{*}{59.90} & \multirow{-2}{*}{\textbf{49.78}} & \multirow{-2}{*}{\textbf{62.34}} & \multirow{-2}{*}{\textbf{74.05}} & \multirow{-2}{*}{\textbf{83.31}} \\
		\hline \noalign{\medskip}
	\end{tabular}
\end{table}

\begin{table}
    \caption{Clustering and recall performance on Cars196~\cite{krause20133d}. Our proposals are highlighted}
	\label{table:results_cars}
	\begin{tabular}{llllll}
		\hline \noalign{\smallskip}
		{Method} & NMI & R@1 & R@2  & R@4 & R@8 \\
		\hline \noalign{\smallskip}
		Semi-hard~\cite{schroff2015facenet} & 53.35 & 51.54 & 63.78 & 73.52 & 82.41 \\[0.25em]
        Lifted Structure~\cite{song2016deep} & 56.88 & 52.98 & 65.70 & 76.01 & 84.27 \\[0.25em]
		N-pairs~\cite{sohn2016improved} & 57.79 & 53.90 & 66.76 & 77.75 & 86.35 \\[0.25em]
		Triplet + Global~\cite{kumar2016learning} & 58.20 & 61.41 & 72.51 & 81.75 & 88.39 \\[0.25em]
		Clustering~\cite{song2016learnable} & 59.04 & 58.11 & 70.64 & 80.27 & 87.81 \\[0.25em]
		\bf{Triplet + FANNG} & 58.24 & 56.11 & 68.34 & 77.99 & 85.92 \\[0.25em]
        \bf{Triplet + FANNG +} \\
        \multicolumn{1}{r}{\bf{Global}} & \multirow{-2}{*}{\textbf{59.70}} & \multirow{-2}{*}{64.20} & \multirow{-2}{*}{75.22} & \multirow{-2}{*}{83.24} & \multirow{-2}{*}{88.94} \\
		\bf{Triplet + FANNG +} \\
		\multicolumn{1}{r}{\bf{Global + Adaptive}} & \multirow{-2}{*}{59.50} & \multirow{-2}{*}{\textbf{64.65}} & \multirow{-2}{*}{\textbf{76.20}} & \multirow{-2}{*}{\textbf{84.23}} & \multirow{-2}{*}{\textbf{90.19}} \\
		\hline \noalign{\medskip}
	\end{tabular}
\end{table}

\begin{figure}
\begin{center}
\begin{tabular}{cc}
\includegraphics[width=0.47\linewidth]{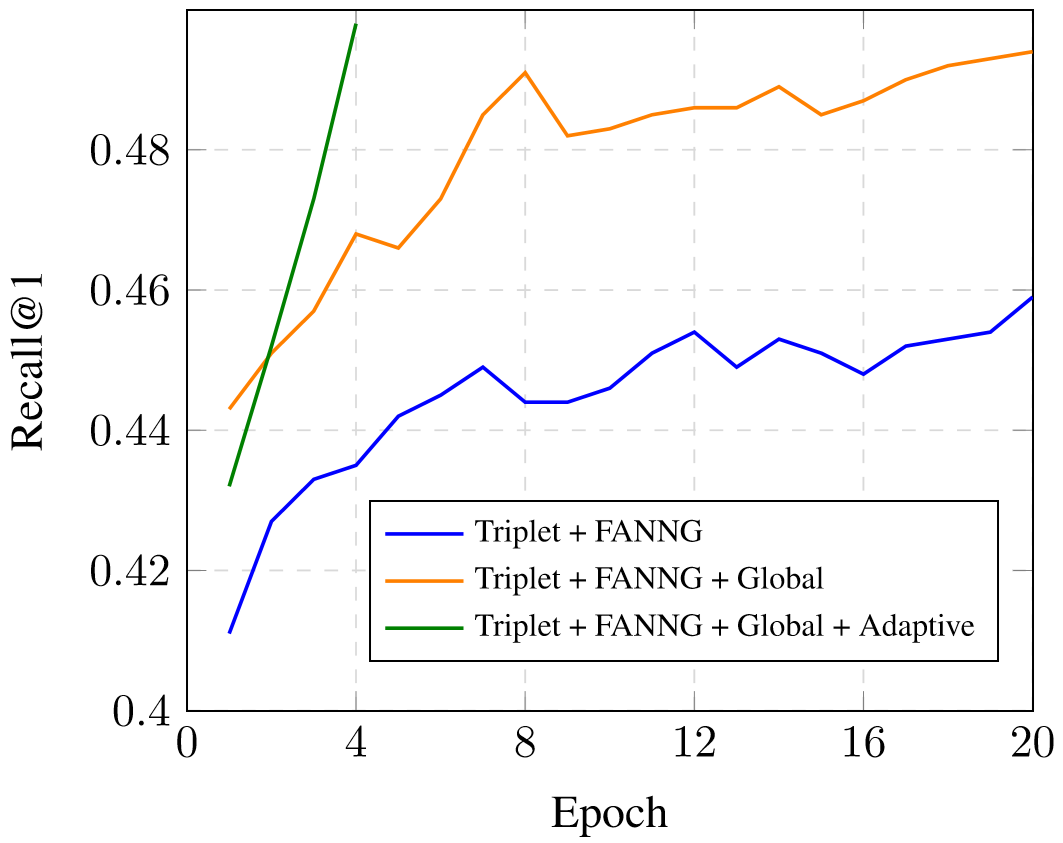} &
\includegraphics[width=0.47\linewidth]{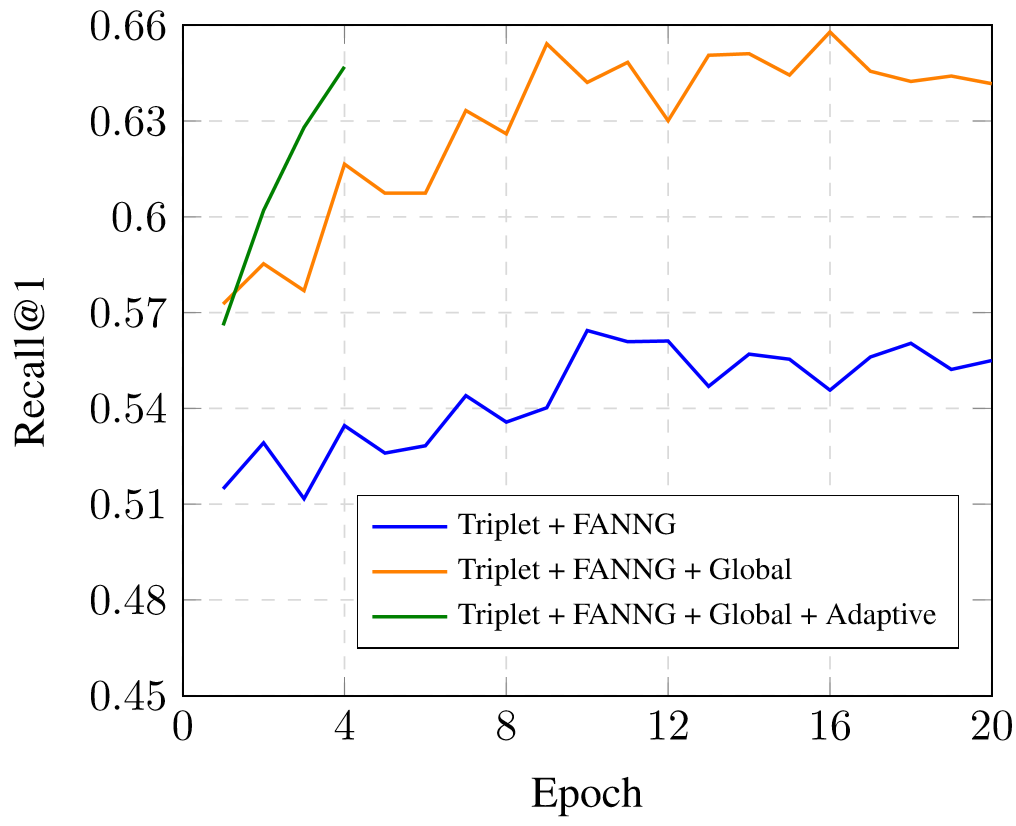} 
\end{tabular}
\end{center}
\caption{A comparison of the convergence rate of our different methods using Recall@1 on CUB-200-2011 dataset (left) and Cars196 (right).}
\label{fig:recall_vs_epochs}
\end{figure}

\subsection{Qualitative Results}

Figures \ref{fig:triplets_birds} and \ref{fig:triplets_cars} 
show triplets for visual inspection. The first column of each figure contains randomly selected anchor points from the training set. Each row then contains the positive and negative sample images that complete each of the triplets. For each of the mined triplets, the negative sample is guaranteed to be a shorter distance from the anchor when compared to the positive sample in the embedding being mined. As per our smart mining algorithm, each of the mined positives is the closest possible positive to the anchor, while still maintaining distance relationships. These properties can be clearly seen when the mined triplets are compared to a randomly generated triplet. The anchors of the mined triplets appear to have stronger similarities with the negative samples, while the random triplet anchors are much closer in appearance to the positive samples. While the mined positive samples are dissimilar from the anchors, in many cases it appears that they are still sharing more features with the anchor than the random positives are sharing with the same anchor. By presenting difficult (but not impossible) triplets more often, our mined triplets enable faster learning of the embedding.

\begin{figure}
	\centering
	\includegraphics[width=0.75\linewidth]{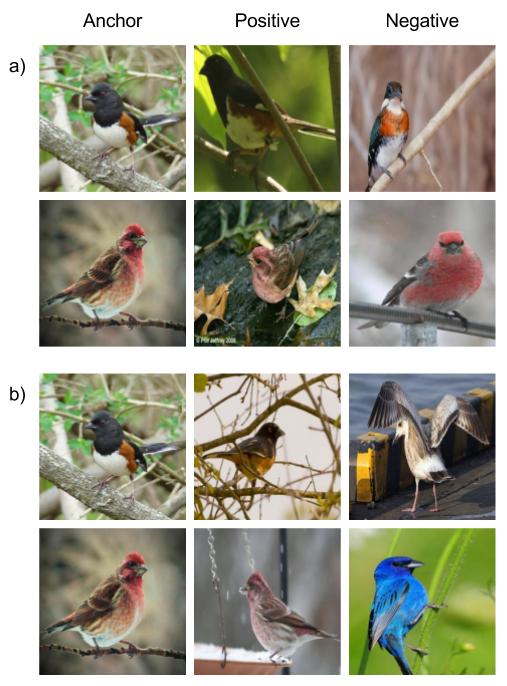} 
	\caption{ a) Triplets mined from the CUB-200-2011~\cite{wah2011caltech} training set using FANNG~\cite{harwood2016fanng}. b) Random triplets constructed with the same anchor. }
	\label{fig:triplets_birds}
\end{figure}
\begin{figure}
	\centering
	\includegraphics[width=0.75\linewidth]{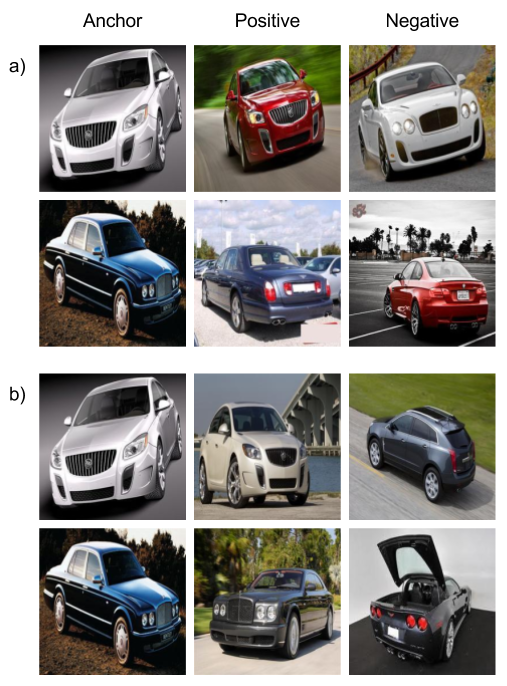} 
	\caption{ a) Triplets mined from the Cars196~\cite{krause20133d} training set using FANNG~\cite{harwood2016fanng}. b) Random triplets constructed with the same anchor. }
	\label{fig:triplets_cars}
\end{figure}


\section{Conclusion}

From the results in Tables~\ref{table:results_cub}-\ref{table:results_cars}, we see that {\bf Triplet + FANNG + Global + Adaptive} significantly outperforms the current state of the art methods~\cite{kumar2016learning,song2016learnable} in terms of clustering and recall performances.  Furthermore, it is worth noting that {\bf Triplet + FANNG} performs substantially better than its counterpart {\bf Semi-hard}~\cite{schroff2015facenet} with respect to the clustering and recall performances, thus highlighting the importance of the smart mining process.  Comparing {\bf Triple + FANNG + Global} and {\bf Triple + FANNG}, we can conclude that the global loss is indeed an important component in improving the clustering and recall performance of the embedding.  Finally, {\bf Triplet + FANNG + Global + Adaptive} and {\bf Triplet + FANNG + Global} show almost equally strong results, but the former has a significantly faster training process.

In this paper, we proposed a novel triplet-based deep metric learning approach that combines a global structure loss with a triplet loss.  We rely on a smart mining process to train our approach, which allows an effective selection of training samples at a low computational cost. Furthermore, we also extend this smart mining with an adaptive controller that automatically selects its hyper-parameters throughout training. By searching the entire training set, we pay a high upfront cost, but make good use of the extra available information to ultimately improve the convergence rate of the training process without compromising on the quality of the embedding.
Using CUB-200-2011~\cite{wah2011caltech} and Cars196~\cite{krause20133d}, we show that our proposed method achieves fast and more accurate training of triplet ConvNets than other competing mining methods. Our approach sets new state-of-the-art deep metric learning results for these two datasets. 
\\\\
\textbf{Acknowledgements:} This research was supported by the Australian Research Council through the Centre of Excellence in Robotic Vision, CE140100016, and through Laureate Fellowship FL130100102 to IDR. We would like thank  Guosheng Lin and Chunhua Shen for their insightful discussions.
{\small
\bibliographystyle{ieee}
\bibliography{tripletFANNG}

\begin{thebibliography}{10}\itemsep=-1pt

\bibitem{bromley1993signature}
J.~Bromley, J.~W. Bentz, L.~Bottou, I.~Guyon, Y.~LeCun, C.~Moore,
  E.~S{\"a}ckinger, and R.~Shah.
\newblock Signature verification using a “siamese” time delay neural
  network.
\newblock {\em International Journal of Pattern Recognition and Artificial
  Intelligence}, 7(04):669--688, 1993.

\bibitem{dosovitskiy2014discriminative}
A.~Dosovitskiy, J.~T. Springenberg, M.~Riedmiller, and T.~Brox.
\newblock Discriminative unsupervised feature learning with convolutional
  neural networks.
\newblock In {\em Advances in Neural Information Processing Systems}, pages
  766--774, 2014.

\bibitem{guillaumin2010multiple}
M.~Guillaumin, J.~Verbeek, and C.~Schmid.
\newblock Multiple instance metric learning from automatically labeled bags of
  faces.
\newblock In {\em Computer Vision--ECCV 2010}, pages 634--647. Springer, 2010.

\bibitem{han2015matchnet}
X.~Han, T.~Leung, Y.~Jia, R.~Sukthankar, and A.~C. Berg.
\newblock Matchnet: Unifying feature and metric learning for patch-based
  matching.
\newblock In {\em Proceedings of the IEEE Conference on Computer Vision and
  Pattern Recognition}, pages 3279--3286, 2015.

\bibitem{harwood2016fanng}
B.~Harwood and T.~Drummond.
\newblock Fanng: Fast approximate nearest neighbour graphs.
\newblock In {\em Proceedings of the IEEE Conference on Computer Vision and
  Pattern Recognition}, pages 5713--5722, 2016.

\bibitem{indefence_HermansBL17}
A.~Hermans, L.~Beyer, and B.~Leibe.
\newblock In defense of the triplet loss for person re-identification.
\newblock {\em CoRR}, http://arxiv.org/abs/1703.07737, 2017.

\bibitem{hoffer2014deep}
E.~Hoffer and N.~Ailon.
\newblock Deep metric learning using triplet network.
\newblock {\em arXiv preprint arXiv:1412.6622}, 2014.

\bibitem{jegou2011product}
H.~Jegou, M.~Douze, and C.~Schmid.
\newblock Product quantization for nearest neighbor search.
\newblock {\em IEEE transactions on pattern analysis and machine intelligence},
  33(1):117--128, 2011.

\bibitem{krause20133d}
J.~Krause, M.~Stark, J.~Deng, and L.~Fei-Fei.
\newblock 3d object representations for fine-grained categorization.
\newblock In {\em Proceedings of the IEEE International Conference on Computer
  Vision Workshops}, pages 554--561, 2013.

\bibitem{kumar2016learning}
B.~Kumar, G.~Carneiro, and I.~Reid.
\newblock Learning local image descriptors with deep siamese and triplet
  convolutional networks by minimising global loss functions.
\newblock {\em Proceedings of the IEEE Conference on Computer Vision and
  Pattern Recognition}, 2016.

\bibitem{manning2008introduction}
C.~D. Manning, P.~Raghavan, H.~Sch{\"u}tze, et~al.
\newblock {\em Introduction to information retrieval}, volume~1.
\newblock Cambridge university press Cambridge, 2008.

\bibitem{masci2014descriptor}
J.~Masci, D.~Migliore, M.~M. Bronstein, and J.~Schmidhuber.
\newblock Descriptor learning for omnidirectional image matching.
\newblock In {\em Registration and Recognition in Images and Videos}, pages
  49--62. Springer, 2014.

\bibitem{Nofuss_Movshovitz-Attias17}
Y.~Movshovitz{-}Attias, A.~Toshev, T.~K. Leung, S.~Ioffe, and S.~Singh.
\newblock No fuss distance metric learning using proxies.
\newblock {\em CoRR}, http://arxiv.org/abs/1703.07464, 2017.

\bibitem{nair2010rectified}
V.~Nair and G.~E. Hinton.
\newblock Rectified linear units improve restricted boltzmann machines.
\newblock In {\em Proceedings of the 27th International Conference on Machine
  Learning (ICML)}, pages 807--814, 2010.

\bibitem{song2016learnable}
H.~Oh~Song, S.~Jegelka, V.~Rathod, and K.~Murphy.
\newblock Deep metric learning via facility location.
\newblock In {\em The IEEE Conference on Computer Vision and Pattern
  Recognition (CVPR)}, July 2017.

\bibitem{schroff2015facenet}
F.~Schroff, D.~Kalenichenko, and J.~Philbin.
\newblock Facenet: A unified embedding for face recognition and clustering.
\newblock In {\em Proceedings of the IEEE Conference on Computer Vision and
  Pattern Recognition}, pages 815--823, 2015.

\bibitem{shrivastava2016training}
A.~Shrivastava, A.~Gupta, and R.~Girshick.
\newblock Training region-based object detectors with online hard example
  mining.
\newblock {\em Proceedings of the IEEE Conference on Computer Vision and
  Pattern Recognition}, 2016.

\bibitem{simo2015discriminative}
E.~Simo-Serra, E.~Trulls, L.~Ferraz, I.~Kokkinos, P.~Fua, and F.~Moreno-Noguer.
\newblock Discriminative learning of deep convolutional feature point
  descriptors.
\newblock In {\em Proceedings of the IEEE International Conference on Computer
  Vision}, pages 118--126, 2015.

\bibitem{Simonyanpami14}
K.~Simonyan, A.~Vedaldi, and A.~Zisserman.
\newblock Learning local feature descriptors using convex optimisation.
\newblock {\em IEEE Transactions on Pattern Analysis and Machine Intelligence},
  2014.

\bibitem{sohn2016improved}
K.~Sohn.
\newblock Improved deep metric learning with multi-class n-pair loss objective.
\newblock In {\em Advances in Neural Information Processing Systems}, pages
  1849--1857, 2016.

\bibitem{song2016deep}
H.~O. Song, Y.~Xiang, S.~Jegelka, and S.~Savarese.
\newblock Deep metric learning via lifted structured feature embedding.
\newblock {\em Proceedings of the IEEE Conference on Computer Vision and
  Pattern Recognition}, 2016.

\bibitem{sung1996learning}
K.-K. Sung.
\newblock Learning and example selection for object and pattern detection.
\newblock 1996.

\bibitem{Szegedy_2015_CVPR}
C.~Szegedy, W.~Liu, Y.~Jia, P.~Sermanet, S.~Reed, D.~Anguelov, D.~Erhan,
  V.~Vanhoucke, and A.~Rabinovich.
\newblock Going deeper with convolutions.
\newblock In {\em The IEEE Conference on Computer Vision and Pattern
  Recognition (CVPR)}, June 2015.

\bibitem{ustinova2016learning}
E.~Ustinova and V.~Lempitsky.
\newblock Learning deep embeddings with histogram loss.
\newblock In {\em Advances in Neural Information Processing Systems}, pages
  4170--4178, 2016.

\bibitem{wah2011caltech}
C.~Wah, S.~Branson, P.~Welinder, P.~Perona, and S.~Belongie.
\newblock The caltech-ucsd birds-200-2011 dataset.
\newblock 2011.

\bibitem{wang2014learning}
J.~Wang, Y.~Song, T.~Leung, C.~Rosenberg, J.~Wang, J.~Philbin, B.~Chen, and
  Y.~Wu.
\newblock Learning fine-grained image similarity with deep ranking.
\newblock In {\em Proceedings of the IEEE Conference on Computer Vision and
  Pattern Recognition}, pages 1386--1393, 2014.

\bibitem{wohlhartcvpr15}
P.~Wohlhart and V.~Lepetit.
\newblock Learning descriptors for object recognition and 3d pose estimation.
\newblock In {\em Proc. IEEE Conference on Computer Vision and Pattern
  Recognition (CVPR)}, 2015.

\bibitem{yuhuihardaware2017}
Y.~Yuan, K.~Yang, and C.~Zhang.
\newblock Hard-aware deeply cascaded embedding.
\newblock {\em CoRR}, http://arxiv.org/abs/1611.05720, 2016.

\bibitem{ZagoruykoCVPR15}
S.~Zagoruyko and N.~Komodakis.
\newblock Learning to compare image patches via convolutional neural networks.
\newblock In {\em Conference on Computer Vision and Pattern Recognition
  (CVPR)}, 2015.

\bibitem{zhuang2016fast}
B.~Zhuang, G.~Lin, C.~Shen, and I.~Reid.
\newblock Fast training of triplet-based deep binary embedding networks.
\newblock {\em Proceedings of the IEEE Conference on Computer Vision and
  Pattern Recognition}, 2016.

\end{thebibliography}
}
\newpage

\section{Effect of Parameters on the Embedding}
In this section, we evaluate the performance of our proposed smart mining method with respect to various parameter settings. Note that in all our experiments (including the ones in the paper), we initialize the network with pre-trained GoogleNet weights \cite{Szegedy_2015_CVPR} and randomly initialize the final fully connected layer similar to \cite{song2016deep} . The learning rate for the randomly initialized fully connected layer is multiplied by 10 to achieve faster convergence similar to \cite{song2016deep}.

\subsection{Effect of Scaling Parameter $\kappa$ on the Embedding}
\label{sec:effect_kappa}
\begin{figure*}
\begin{center}
\begin{tabular}{cc}
\includegraphics[width=0.45\textwidth]{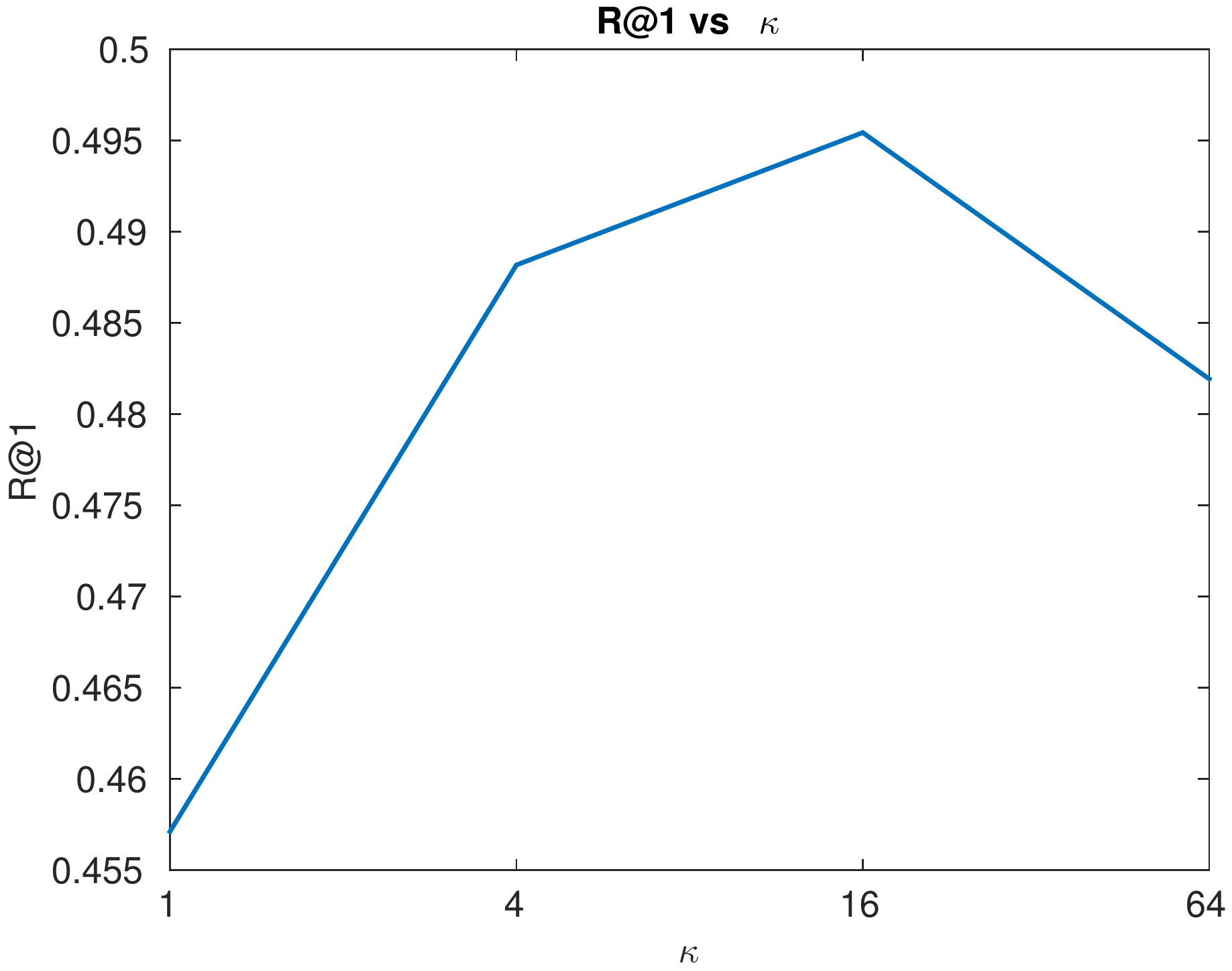} &
\includegraphics[width=0.45\textwidth]{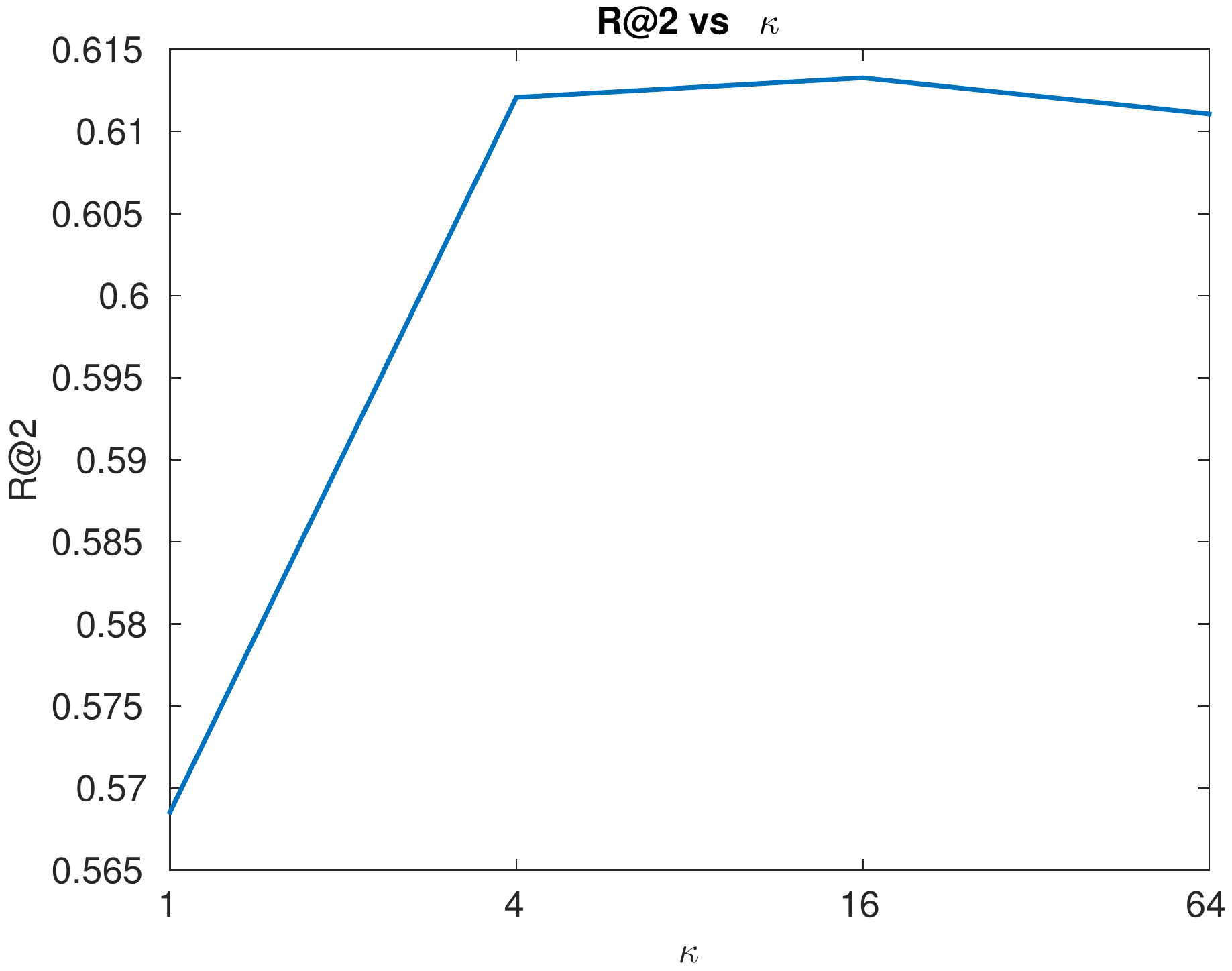} \\
\includegraphics[width=0.45\textwidth]{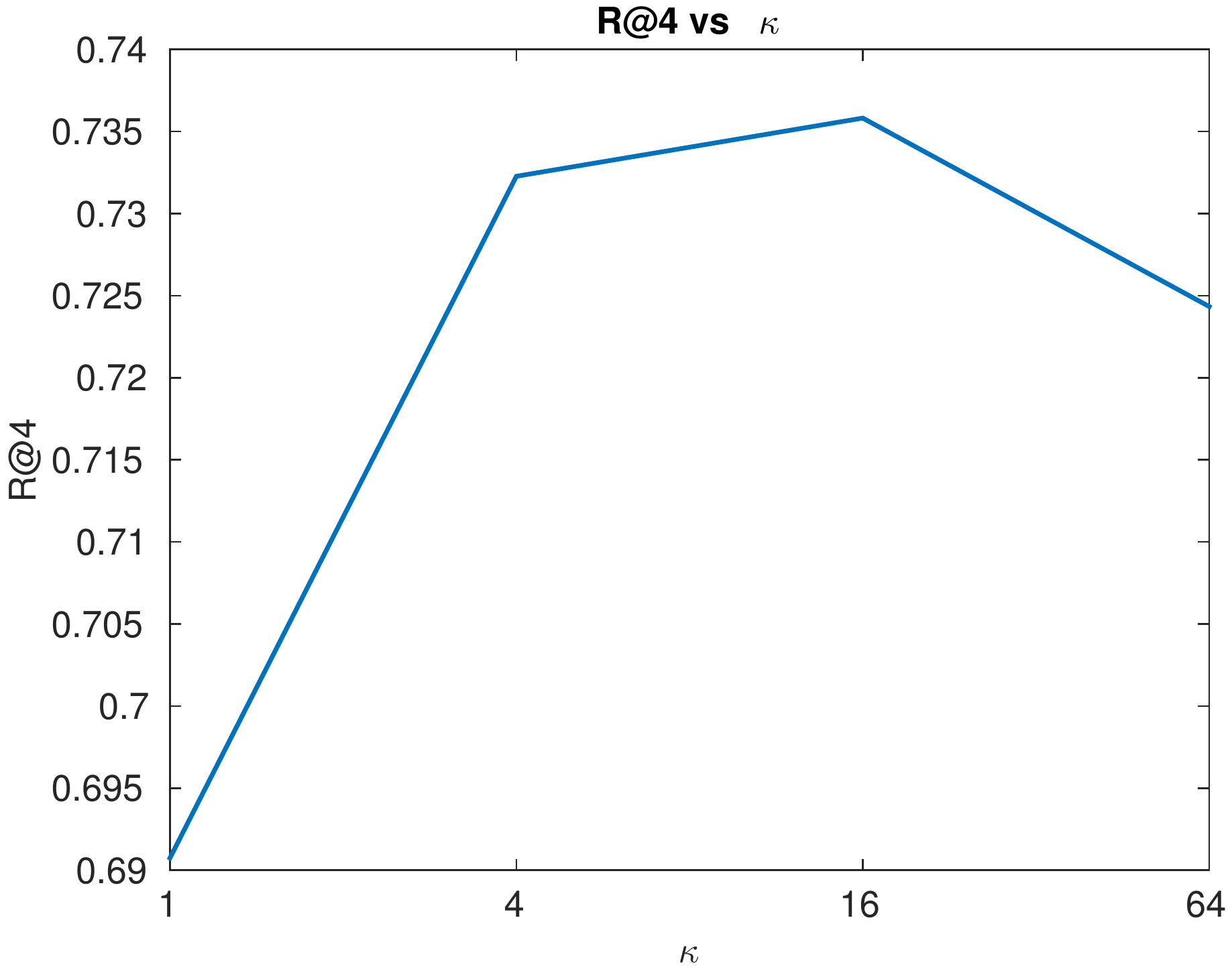} &
\includegraphics[width=0.45\textwidth]{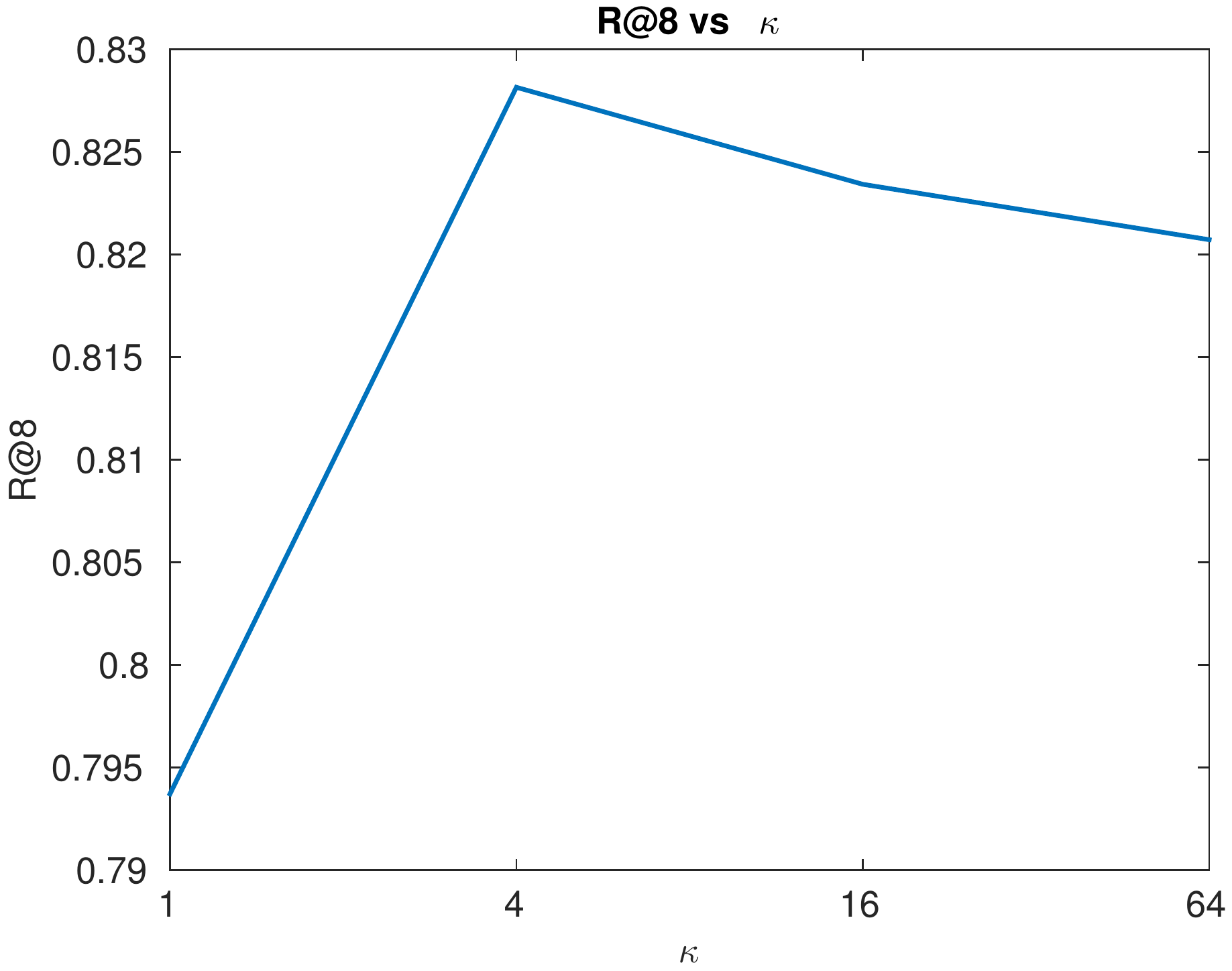} 
\end{tabular} 
\end{center}
\caption{$R@1$ vs $\kappa$ (top left), $R@2$ vs $\kappa$ (top right), $R@4 $vs $\kappa$ (bottom left), $R@8$ vs $\kappa$ (bottom right)}
\label{fig:recall_graphs}
\end{figure*}
\noindent We define smart triplets as those that satisfy Eq. $6$, where $\kappa$ is a global scaling factor that decides the radius of the hyper-spherical exclusion boundary centred around the anchor. In this sub-section, we show the effect of $\kappa$ on the feature embedding. To this end, we run the experiments on CUB-200-2011 dataset for different initial values of $\kappa\in \{ 1,4,16,64\}$. We use \textbf{Triplet + FANNG + Global} as the loss function and report the recall values at $1,2,4$ and $8$ at the end of $20^{th}$ epoch. Fig.~\ref{fig:recall_graphs} shows that the performance degrades for smaller values of $\kappa$. This is due to hard triplets generated by the mining algorithm. For large values of $\kappa$, there are fewer smart triplets returned by the approximate nearest neighbor search, so random triplets are used instead.  In the latter case, the behavior of the method tends to be similar to that of the \textbf{Triplet + Global}.

\subsection{Effect of the Percentage of Mined Triplets for Training}
\label{sec:effect_percent_trip}
\begin{figure*}
\begin{center}
\begin{tabular}{cc}
\includegraphics[width=0.45\textwidth]{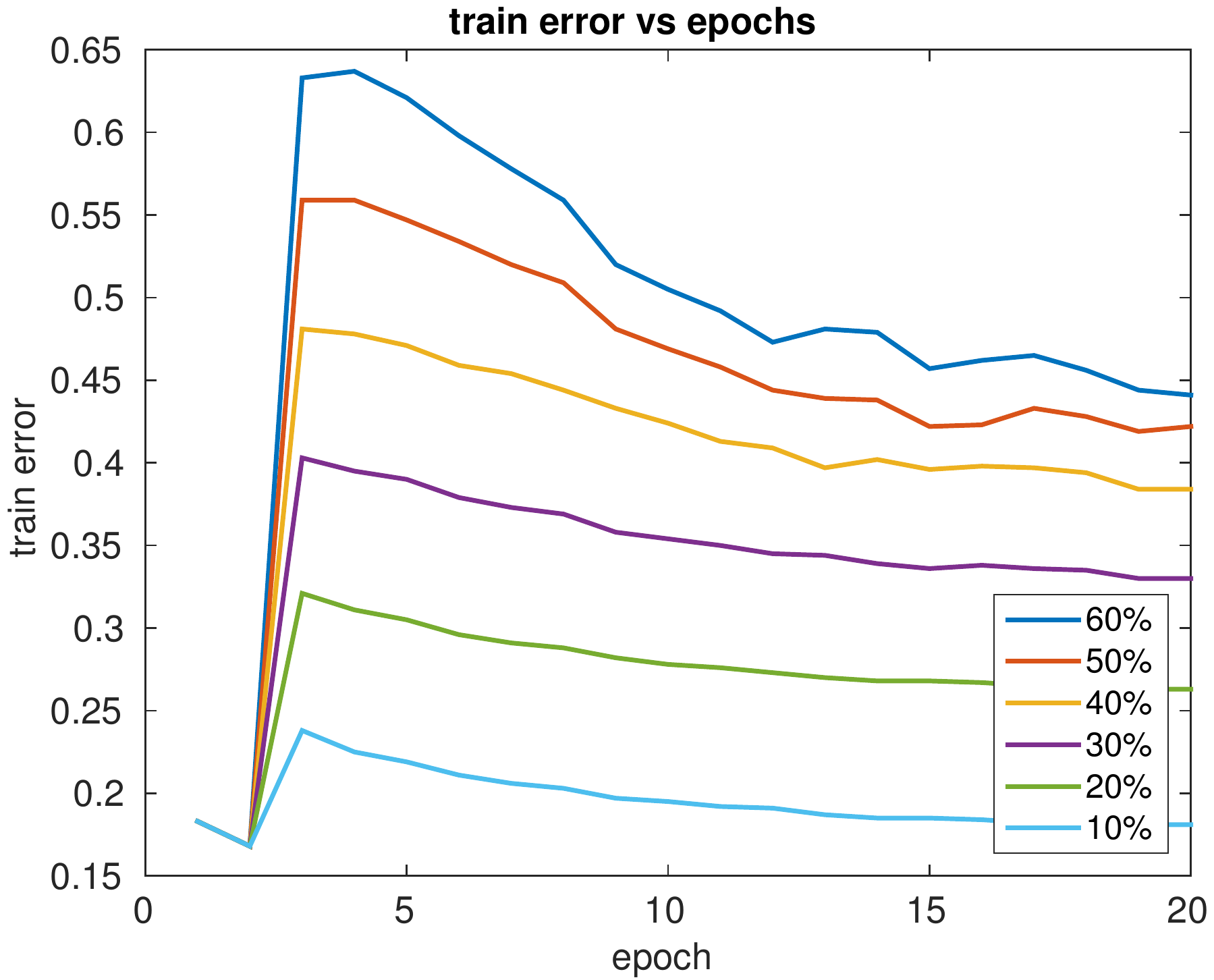} &
\includegraphics[width=0.45\textwidth]{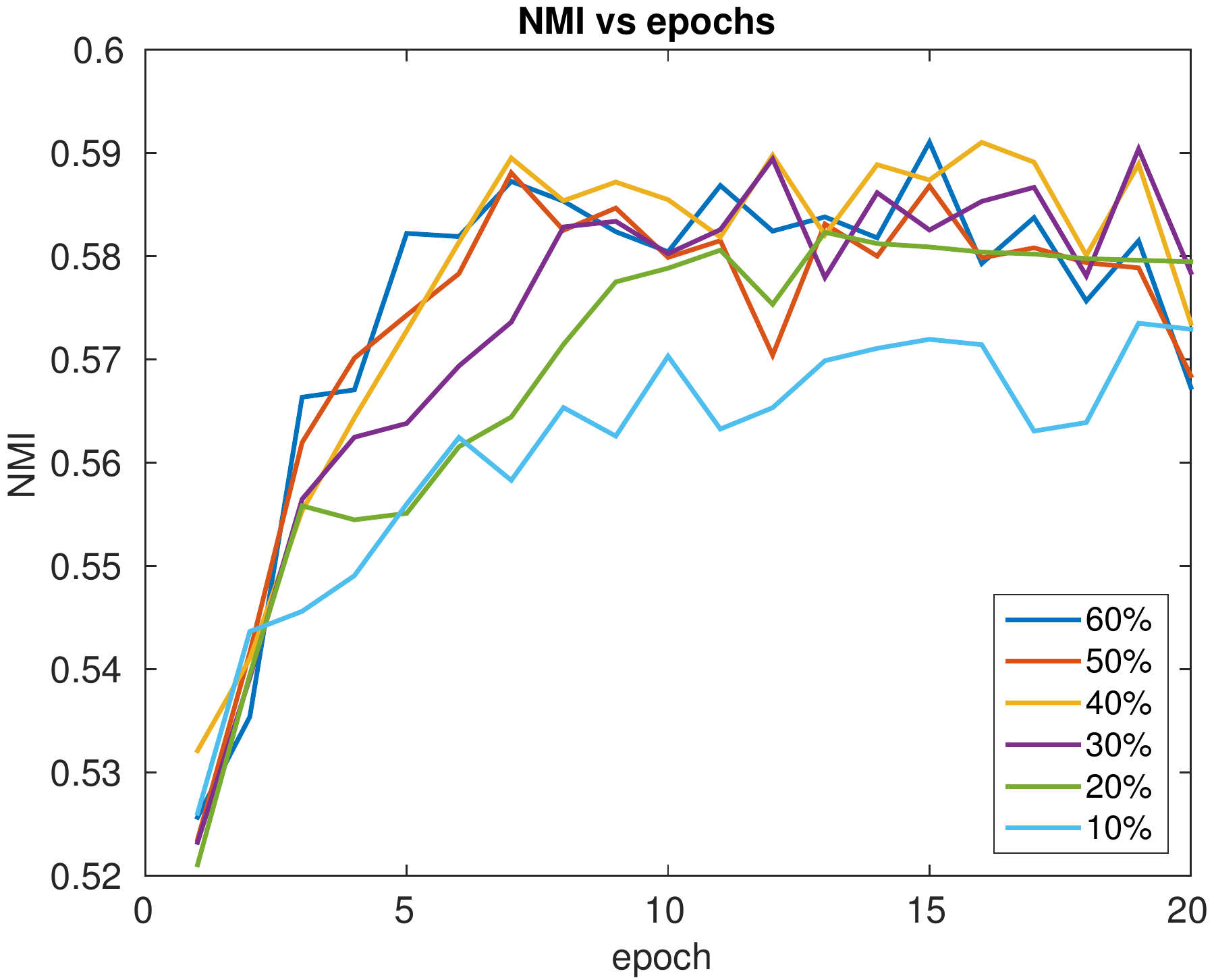} \\
\includegraphics[width=0.45\textwidth]{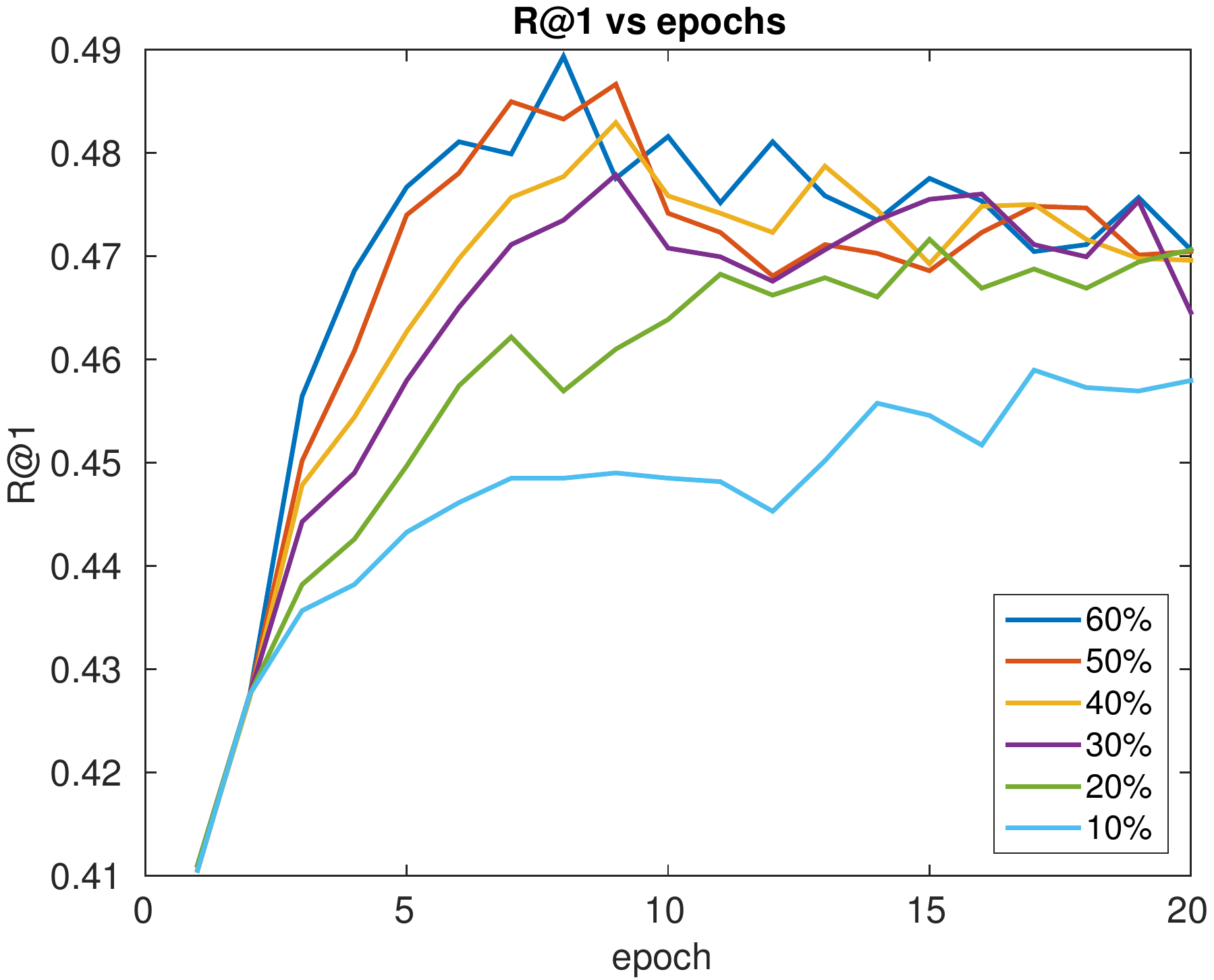} &
\includegraphics[width=0.45\textwidth]{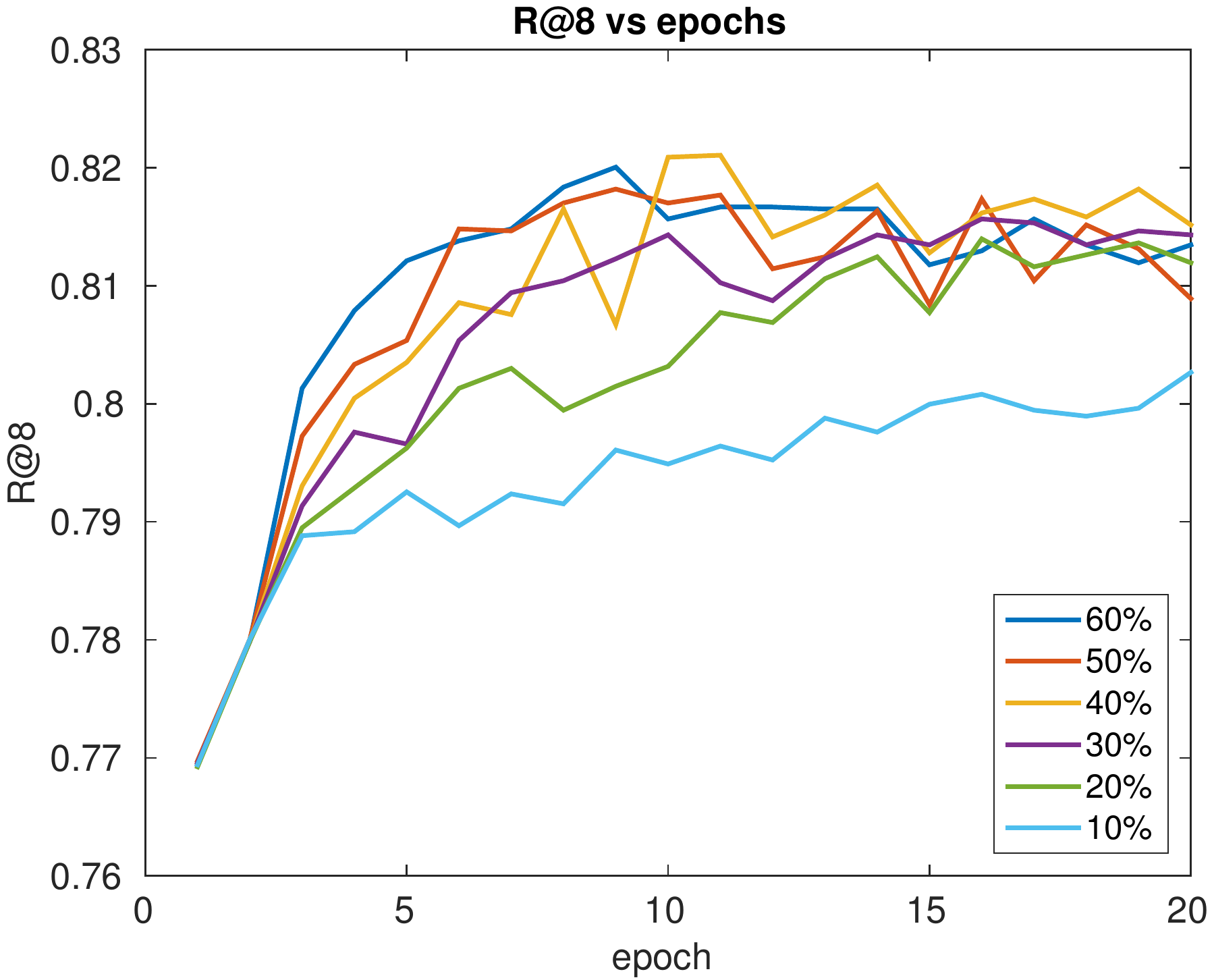} 
\end{tabular} 
\end{center}
\caption{Training error vs epoch (top left), NMI vs epoch (top right), $R@1 $vs epoch (bottom left), $R@8$ vs epoch (bottom right)}
\label{fig:percentage_triplets}
\end{figure*}
Figure \ref{fig:percentage_triplets} shows the effect of varying the percentage of mined triplets for training on the CUB-200-2011 dataset. We train \textbf{Triplet + FANNG + Adaptive} networks for $20$ epochs using a target training error of $0.5$ and with the percentage of mined triplets varying from $10\%$ to $60\%$ in $10\%$ increments. For these experiments the global loss has been disabled so that the training error is a result of only the triplet losses. At the lower percentages, there are insufficient mined triplets to properly control the training error and accelerate the training. From $40\%$ mined triplet and beyond, there are enough mined triplets to allow for control the training error and so the performance begins to saturate at this level. As such, we find that a percentage of anywhere between $50\%$ to $100\%$ mined triplets is sufficient.

\section{Visualizing Embedding using t-SNE}
\label{sec:tsne_cub}
Fig.~\ref{fig:tsen-birds} shows the Barnes-Hut t-SNE visualisation of the learned embedding space obtained by mapping the CUB-200-2011 test image features to a two-dimensional space. Although, there is no overlap between the train and test classes, the images from the test classes are clustered well.
\begin{figure*}
\begin{center}
\includegraphics[width=1\textwidth]{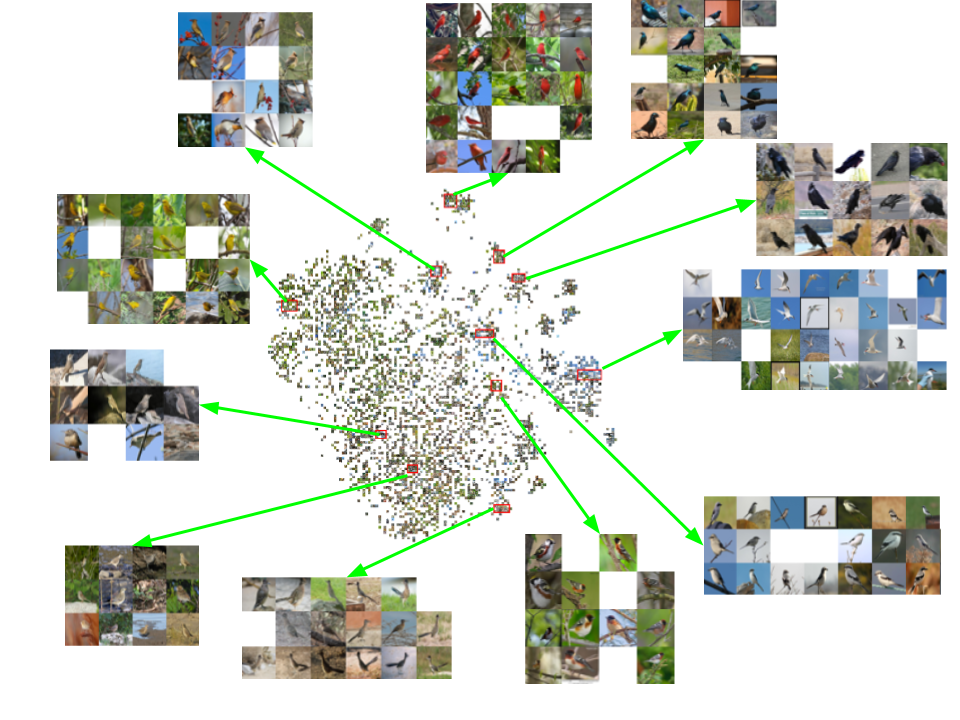}\\ 
\end{center}
\caption{Barnes-Hut t-SNE visualization of the CUB-200-2011 test images}
\label{fig:tsen-birds}
\end{figure*}

\section{Sample Mined Triplets using FANNG}
\label{sec:triplet_cub}
\begin{figure*}
\begin{center}
\begin{tabular}{c}
\includegraphics[width=0.85\textwidth]{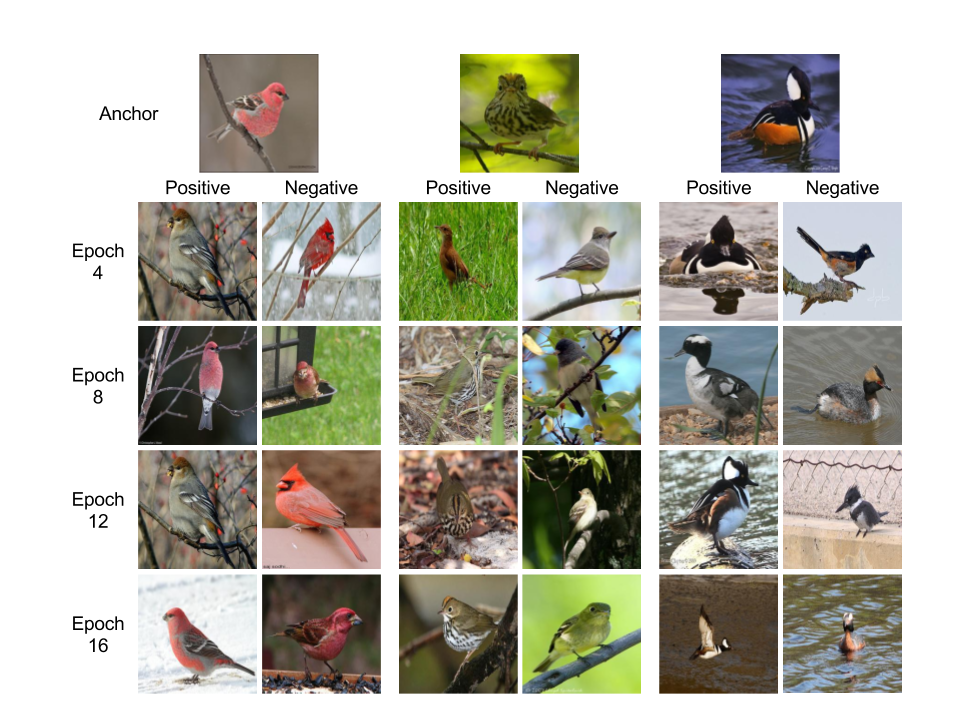}\\ 
\includegraphics[width=0.85\textwidth]{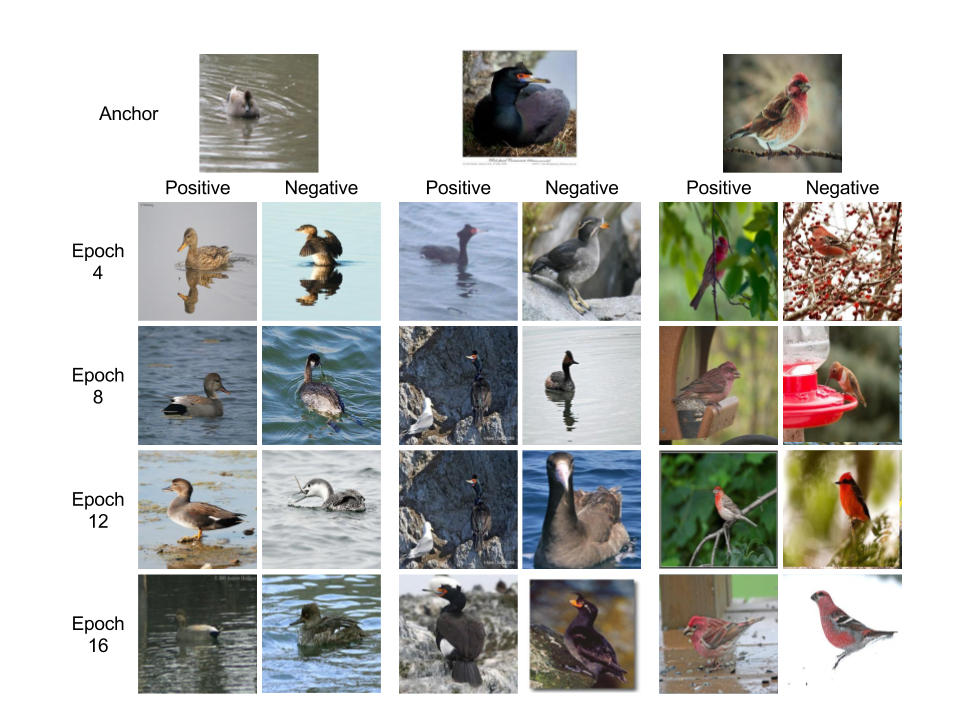}
\end{tabular} 
\end{center}
\caption{Mined triplets for $6$ specific anchor points at training epochs $4,8,12$ and $16$.}
\label{fig:sample_triplets}
\end{figure*}
The images in Figure~\ref{fig:sample_triplets} are triplets from randomly selected anchor points while training \textbf{Triplet + FANNG + Adaptive} on the CUB-200-2011 dataset. Similar to the experiments in Section \ref{sec:effect_percent_trip}, we are interested in showing only the learnning resulting from the triplet mining and as such global loss is disabled. At epochs $4,8,12$ and $16$ the first triplet formed for each of the chosen anchor points was recorded. Beginning with the epoch $4$ images, visual inspection shows that the mined negative samples share distinct visual traits with the anchor image and hence they are already much harder than random negatives. Beyond epoch $4$, the mined negatives continue to become more difficult as the embedding is refined. In particular, many of the negative images at Epoch $16$ could easily be mistaken as coming from the same class as the anchor image. The appearence of the positive samples is largly constrained by the negatives, since our method always selects the softest positive that is also still harder that the chosen negative. This selection process can be seen in the way each positive-negative pair share many distinctive visual traits such that they are roughly the same distance from the anchor point. However, in some cases the negative and positive samples could be in very different directions from the anchor, and so visually judging the similar level of difficulty is much more difficult across different regions of the embedding.

\end{document}